\documentclass{article}

\usepackage[preprint]{neurips_2025}

\usepackage[utf8]{inputenc} 
\usepackage[T1]{fontenc}    
\usepackage{hyperref}       
\usepackage{url}            
\usepackage{booktabs}       
\usepackage{amsfonts}       
\usepackage{nicefrac}       
\usepackage{microtype}      
\usepackage{xcolor}         
\usepackage{amsmath}        
\usepackage{amsmath}  
\usepackage{amssymb, bbm}

\usepackage{multirow}       
\usepackage{bm}
\usepackage{graphicx}   
\usepackage{pgfplots}
\usepackage{adjustbox}
\usepackage{caption}
\usepackage{subcaption}
\usepackage{pgfplots}
\usepackage{pgfplotstable}
\usepackage{tikz}
\usepackage{caption}
\usepackage{subcaption}
\pgfplotsset{compat=1.18}  
\usetikzlibrary{pgfplots.groupplots}
\usepackage[linesnumbered,ruled,vlined]{algorithm2e}  
\usepackage{amsmath}     
\usepackage{amssymb}     
\usepackage{bbm}         
\usepackage{xcolor}
\usepackage{wrapfig}

\usepackage[ruled,vlined]{algorithm2e}
\SetKwInput{Input}{Input}
\usepackage{float} 

\definecolor{steelblue}{RGB}{70,130,180}
\definecolor{crimson}{RGB}{220,20,60}
\definecolor{sienna}{RGB}{160,82,45}
\definecolor{goldenrod}{RGB}{218,165,32}
\newtheorem{theorem}{Theorem}

\usepackage{marginnote}

\title{TRIX- Trading Adversarial Fairness via Mixed Adversarial Training}

\author{%
  Tejaswini Medi \\
  University of Mannheim \\
  \texttt{tejaswini.medi@uni-mannheim.de} \\
  \And
  Steffen Jung \\
  University of Mannheim \\
  MPI for Informatics \\
  \And
  Margret Keuper \\
  University of Mannheim \\
  MPI for Informatics \\
}

\author{Tejaswini Medi$^{1}$, Steffen Jung$^{1,2}$, Margret Keuper$^{1,2}$ \\
\small $^1$University of Mannheim, Germany $^2$MPI for Informatics, Saarland Informatics Campus, Germany \\
\small tejaswini.medi@uni-mannheim.de
}

\begin{document}

\maketitle

\begin{abstract}
Adversarial Training (AT) is a widely adopted defense against adversarial examples. However, existing approaches typically apply a uniform training objective across all classes, overlooking disparities in class-wise vulnerability. This results in \textit{adversarial unfairness}: classes with well distinguishable features (\textit{strong classes}) tend to become more robust, while classes with overlapping or shared features (\textit{weak classes}) remain disproportionately susceptible to adversarial attacks. We observe that strong classes do not require strong adversaries during training, as their non-robust features are quickly suppressed. In contrast, weak classes benefit from stronger adversaries to effectively reduce their vulnerabilities. Motivated by this, we introduce \textbf{TRIX}, a feature-aware adversarial training framework that adaptively assigns weaker \textit{targeted} adversaries to strong classes, promoting feature diversity via uniformly sampled targets, and stronger \textit{untargeted} adversaries to weak classes, enhancing their focused robustness. TRIX further incorporates per-class loss weighting and perturbation strength adjustments, building on prior work, to emphasize weak classes during the optimization. Comprehensive experiments on standard image classification benchmarks, including evaluations under strong attacks such as PGD and AutoAttack, demonstrate that TRIX significantly improves worst-case class accuracy on both clean and adversarial data, reducing inter-class robustness disparities, and preserves overall accuracy. Our results highlight TRIX as a practical step toward fair and effective adversarial defense.
\end{abstract}

\section{Introduction} \label{Intro}
Deep neural networks (DNNs) have achieved remarkable success across a wide range of domains, including computer vision~\citep{he2016deep, dosovitskiy2021image}, natural language processing~\citep{devlin2019bert, brown2020language}, and decision-making~\citep{silver2017mastering}, often surpassing human-level performance. Despite these achievements, DNNs remain vulnerable to \emph{adversarial examples}—carefully crafted, imperceptible perturbations that can induce confident misclassifications~\citep{szegedy2014intriguing, goodfellow2014explaining, madry2018towards}. Such vulnerabilities pose significant risks in safety-critical applications like autonomous driving and medical diagnostics~\citep{kurakin2017adversarial, carlini2017towards}, where even minor errors can lead to severe consequences. As a result, various defense mechanisms have been proposed to improve the robustness of deep neural networks. Among them, \emph{Adversarial Training} (AT) has emerged as one of the most effective defenses~\citep{agnihotri2025sembench, athalye2018obfuscated, Wang2020ImprovingAR, jia2022prior, 10.1007/978-3-031-73636-0_21, agnihotri2023cospgd, agnihotri2023unreasonable, grabinski2024large, Grabinskilowcut22, grabinski2022aliasing, grabinski2022robust, Jung2023, lukasik2023improving}. Notably, TRADES~\citep{zhang2019theoretically} provides a principled AT framework that balances clean accuracy with adversarial robustness by minimizing a combination of the standard cross-entropy loss and a Kullback–Leibler divergence between model predictions on clean and adversarial examples.

\begin{figure}[ht]
\centering
\begin{minipage}[t]{0.5\textwidth}
    \centering
        \scriptsize
    \begin{tikzpicture}
    \begin{axis}[
    width=\textwidth,
    height=3.5cm,
        ybar,
        bar width=4pt,
        ymin=0, ymax=100,
        ylabel={ASR (\%)},
        symbolic x coords={airplane, automobile, bird, cat, deer, dog, frog, horse, ship, truck},
        xtick=data,
        x tick label style={rotate=45, anchor=east},
        enlarge x limits=0.15,
        bar shift=0pt,
        legend style={
            at={(0.5,1.05)},
            anchor=south,
            legend columns=2,
            font=\scriptsize,
            draw=none,
            /tikz/every even column/.append style={column sep=0.4cm}
        },
    ]
    \addplot+[] coordinates {
        (airplane,34.9) (automobile,25.5) (bird,65.7) (cat,73.7) (deer,64.5)
        (dog,57.2) (frog,41.4) (horse,36.7) (ship,30.0) (truck,31.2)
    };
     \addplot+[] coordinates {
        (airplane,20.1) (automobile,14.1) (bird,40.3) (cat,40.3) (deer,48.8)
        (dog,37.9) (frog,24.2) (horse,21.7) (ship,18.0) (truck,18.4)
    };
    \addplot+[] coordinates {
        (airplane,7.7) (automobile,4.7) (bird,11.9) (cat,14.2) (deer,13.8)
        (dog,12.4) (frog,13.5) (horse,8.3) (ship,7.7) (truck,7.7)
    };


    \legend{Untargeted ASR, Untargeted ASR (KL), Targeted ASR (mean) }

    \end{axis}
    \end{tikzpicture}
    \caption*{(a) Attack Success Rates}
\end{minipage}
\hfill
\begin{minipage}[t]{0.48\textwidth}
    \centering
    \scriptsize
    \begin{tikzpicture}
    \begin{axis}[
    width=\textwidth,
    height=3.5cm,
        ybar,
        bar width=4pt,
        ymin=0, ymax=1.0,
        ylabel={Disparity Value},
        symbolic x coords={Std Dev,\phantom{ccc}Min-Max,Variance,Avg Acc,Min Acc},
        xtick=data,
        x tick label style={rotate=45, anchor=east},
        enlarge x limits=0.15,
        bar shift auto,
        legend style={
            at={(0.5,1.05)},
            anchor=south,
            legend columns=2,
            font=\scriptsize,
            draw=none,
            /tikz/every even column/.append style={column sep=0.4cm}
        }
    ]

    \addplot+[] coordinates {
        (Std Dev,0.0934) (\phantom{ccc}Min-Max,0.258) (Variance,0.1133) (Avg Acc,0.8241) (Min Acc,0.668)
    };
    \addplot+[] coordinates {
        (Std Dev,0.0752) (\phantom{ccc}Min-Max,0.217) (Variance,0.0913) (Avg Acc,0.8233) (Min Acc,0.707)
    };
    \addplot+[] coordinates {
        (Std Dev,0.1583) (\phantom{ccc}Min-Max,0.452) (Variance,0.2682) (Avg Acc,0.5903) (Min Acc,0.326)
    };
    \addplot+[] coordinates {
        (Std Dev,0.1206) (\phantom{ccc}Min-Max,0.32) (Variance,0.2085) (Avg Acc,0.5783) (Min Acc,0.424)
    };

    \legend{
        Clean - TRADES,
        Clean - Our Model,
        Robust - TRADES,
        Robust - Our Model
    }

    \end{axis}
    \end{tikzpicture}
    \caption*{(b) Disparity metrics and accuracies}
\end{minipage}
\hfill
\caption{
\textbf{(a)} Class-wise Attack Success Rates (ASR) under PGD attack with epsilon (8/255) for both untargeted and targeted attacks. Weak classes such as \textit{cat} and \textit{bird} exhibit higher vulnerability, reflected in elevated ASR values. Targeted attacks, however weaker than the untargeted attacks.
\textbf{(b)} Comparison of disparity metrics and accuracy between TRADES and our model. Our method demonstrates lower disparity (Std Dev, Min-Max, Variance) and improved minimum robust accuracy (Min Acc), indicating better fairness and robustness balance.
}
\label{fig:comparison_plots}
\end{figure}

However, a critical limitation of TRADES and other standard AT defense methods is their assumption of uniform vulnerability across all classes. These methods apply the same adversarial objective to every class, overlooking substantial differences in class-wise robustness and their corresponding intrinsic feature diversity and shared feature representations. In practice, certain \emph{strong} classes exhibit well-separated and stable features, while others, termed \emph{weak} classes, suffer from overlapping or fragile feature representations that make them more vulnerable to adversarial perturbations~\citep{tian2021analysis, xu2021robust}. This disparity in class-wise robustness performance, known as the \emph{Adversarial Fairness Problem}~\citep{xu2021robust, zhang2021dafa, medi2024classwiserobustnessanalysis, medi2025fair, wei2023cfa, fairness_bat,li2023wat, fairness_frl}, results in disproportionate degradation of weak or minority class performance under adversarial conditions, raising fairness concerns in real-world deployment~\citep{ma2021understanding}.

To address this issue, recent methods have proposed class-aware adversarial training strategies, such as class-wise reweighting~\citep{lee2024dafa, ma2022tradeoff, zhao2023improving,wei2023cfa} and the inclusion of targeted perturbations based on class confusion~\citep{medi2025fair}. For example, FAIR-TAT~\citep{medi2025fair} constructs targeted adversaries guided by class-to-class confusion patterns in AT, aiming to improve the separation of entangled decision boundaries. DAFA~\citep{lee2024dafa}, a recent work introduces inter-class feature similarity distance as a dynamic signal to adaptively adjust per-class loss weights and perturbation margins. While these approaches enhance robust fairness, they often come at the cost of clean-sample fairness or overall accuracy. A fair and robust model should maintain clean and robust sample fairness while maintaining the overall accuracies on both.

\textbf{In this work, we ask: \emph{Can we achieve better class-wise performance in fairness on clean and robust samples by tailoring the type of adversary to each class’s feature characteristics and vulnerability?}}

While prior works have primarily focused on either the class-wise perturbation strength of untargeted adversaries or targeted adversaries with respect to strong and weak classes, our work takes a step further by analyzing the \textit{type} of adversary best suited for each class within adversarial training regime. 
Although previous studies suggest that strong classes benefit from well-separated features, our analysis also reveals a more nuanced behavior: Strong classes due to their structured and semantically aligned representations enable them to suppress non-robust features more effectively by minimizing the loss early during training \citep{li2024adversarial}. In contrast, weak classes struggle to eliminate non-robust components, even as training progresses due to their intrinsic feature diversity. Conversely, while existing fairness-driven methods improve performance for weaker classes by emphasizing them during training, they often overlook a critical limitation: these classes typically exhibit high intra-class variability, with latent representations that lack sufficient separability.

\begin{figure}
    \centering
    \includegraphics[width=\textwidth]{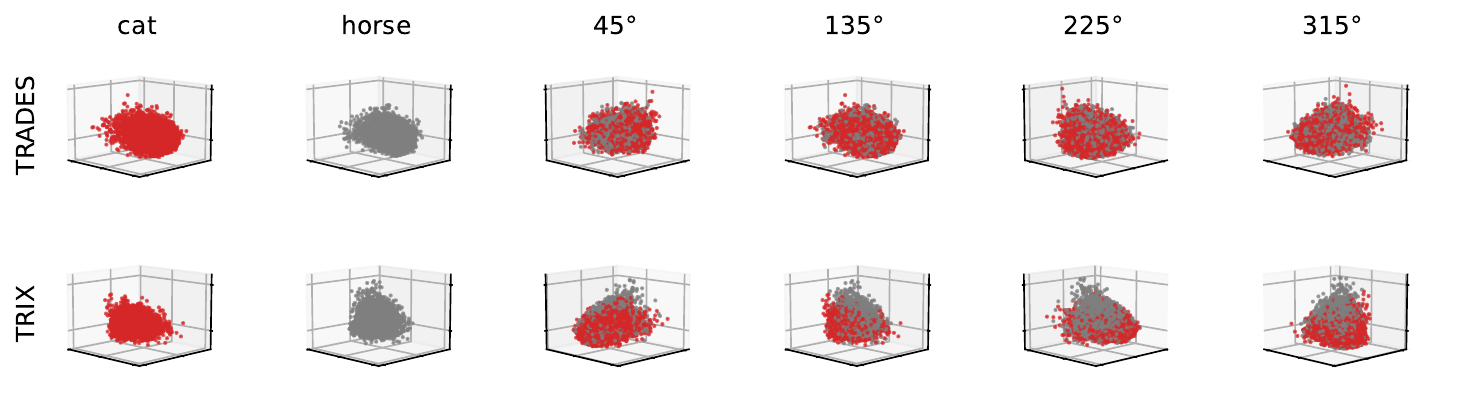}
    \caption{%
    Features of classes \textit{cat} (first column) and \textit{horse} (second column) attacked with PGD and projected to the first three principal components.
    We show both classes combined at different angles in the remaining four columns.
    TRADES shows more overlap, indicating higher confusion under perturbation.
    These visualizations illustrate stronger class-wise feature compactness achieved by TRIX.
    }
    \label{fig:cat-horse}
\end{figure}

To address these limitations in existing adversarial training (AT)-based fairness approaches, we propose a novel \textit{feature-aware adversarial training} strategy that dynamically modulates the nature of adversarial perturbations based on class-wise performance. Specifically, we apply \textit{stronger untargeted adversarial perturbations} to underperforming (weak) classes in order to enhance their robustness by maximally expanding their decision boundaries, leveraging their intrinsic diversity. In contrast, we apply \textit{weaker targeted perturbations} to strong classes to encourage feature entanglement across class boundaries and prevent dominance. This design encourages richer, more robust representations for weak classes while preserving the structured separability of strong classes. Crucially, by \textit{uniformly sampling target labels} during targeted attacks, our method increases class-to-class diversity and enables the model to better capture inter-class bias relationships.


Fig.~\ref{fig:comparison_plots}(a) shows that untargeted attacks generally achieve higher success rates than targeted ones, even on an adversarially trained model using TRADES.
Building on this, our class-aware adversarial training scheme shifts adversarial focus toward vulnerable regions of the feature space, using strength variations in varied adversaries, improving robustness where it is most needed.
We show the discriminative capability of TRIX in comparison with TRADES considering two classes from CIFAR10:
\textit{cat} (weakest animal class) and \textit{horse} (strong animal class)    using first three principal component feature projections.
As illustrated in Fig.~\ref{fig:cat-horse}, a TRADES-trained robust model shows more overlap between \textit{cat} and \textit{horse}.
In contrast, TRIX shows reduced overlap, thus demonstrates improved discriminative capability.
Overall, our approach not only enhances robustness for weak classes but also encourages broader and fairer class separation in the latent space, resulting in more reliable predictions (refer Fig.~\ref{fig:comparison_plots}(b)).

\textbf{We term our approach \emph{TRIX} (TRades-based mIXed adversarial training)}, a novel framework that combines targeted and untargeted adversaries in a principled, class-specific manner. 


\textbf{Our contributions are summarized as follows:}
\begin{itemize}
    \item We identify and empirically validate the \textit{asymmetric suppression} of non-robust features across classes. Specifically, we observe that \textbf{weaker classes} are more severely affected by stronger adversarial perturbations, leading to a notable drop in performance, whereas \textbf{stronger classes} exhibit greater resilience, maintaining their performance even under aggressive attacks.

    \item We also empirically analyze the difference between targeted and untargeted adversarial training to study the influence of type of adversaries on class-wise robustness.

    \item We propose \textbf{TRIX}, a fairness-aware adversarial training method that dynamically assigns either targeted or untargeted adversaries in AT based on class-wise feature similarity distance. TRIX also uses similarity-driven loss weighting to emphasize weak classes during optimization, following prior work.

    \item Experiments on CIFAR-10, CIFAR-100, and STL-10 show that TRIX significantly improves worst-class accuracy and inter-class fairness on both clean and robust samples, while maintaining respective overall accuracies under strong attacks, like AutoAttack and PGD.
\end{itemize}

\section{Preliminaries}

\paragraph{Adversarial Training with Targeted and Untargeted Objectives.}
In standard adversarial training, models are optimized using inputs perturbed by \emph{untargeted} adversarial attacks. These attacks aim to increase the overall classification loss and induce misclassification toward any class other than the ground-truth label. This formulation enhances robustness by encouraging the model to maintain correct predictions despite perturbations within a bounded $\ell_p$-ball~\citep{madry2018towards}. 

An alternative and complementary strategy is \emph{targeted adversarial training}, where the adversarial perturbation is constructed to explicitly push the model's prediction toward a specific incorrect target class~\citep{targeted_adv_training}. The target class $\tilde{y} \neq y$ can be selected either uniformly at random or guided by specific objectives, such as targeting the most confusing classes—as done in FAIR-TAT~\citep{medi2025fair}. Targeted adversarial training can be particularly beneficial in scenarios involving class imbalance or overconfidence, as it directly regularizes the margin between the correct class and a chosen target, thereby shaping more discriminative boundaries.

We extend the TRADES framework~\citep{zhang2019theoretically}, which balances natural accuracy and robustness, to support both \textit{targeted} and \textit{untargeted} adversaries. 
The general TRADES objective is formulated as:

\begin{equation}
\label{eq:trades}
\min_{\theta} \; \mathbb{E}_{(\bm{x}, y) \sim \mathcal{D}} \left[
    \mathcal{L}_{\mathrm{CE}}(f_\theta(\bm{x}), y) +
    \beta \cdot \text{KL}(f_\theta(\bm{x}) \| f_\theta(\bm{x}^{\mathrm{adv}}))
\right],
\end{equation}

where $f_\theta$ denotes the model with parameters $\theta$, $\mathcal{L}_{\mathrm{CE}}$ is the standard cross-entropy loss, and $\bm{x}^{\mathrm{adv}}$ is an adversarial example constrained within an $\epsilon$-ball around the original input $\bm{x}$.

In the \textbf{untargeted} setting, the adversary aims to maximize the KL divergence between the model's prediction on clean input and that on the perturbed input:

\begin{equation}
\label{eq:untargeted}
\bm{x}^{\mathrm{adv}} = \arg\max_{\|\bm{\delta}\| \leq \epsilon} \; \text{KL}(f_\theta(\bm{x}), f_\theta(\bm{x} + \bm{\delta})).
\end{equation}

In contrast, the \textbf{targeted} setting selects a specific incorrect class $\tilde{y} \neq y$ and seeks perturbations that minimize the cross-entropy loss toward this target:

\begin{equation}
\label{eq:targeted}
\bm{x}^{\mathrm{adv}}_{t} = \arg\min_{\|\bm{\delta}\| \leq \epsilon} \; \mathcal{L}_{\mathrm{CE}}(f_\theta(\bm{x} + \bm{\delta}), \tilde{y}).
\end{equation}

To unify these two paradigms, we propose a mixed adversarial objective, where the type of attack applied to each sample is determined by a class based feature-conditional policy $\pi(y) \in \{\text{targeted}, \text{untargeted}\}$. This allows the model to adaptively apply targeted or untargeted attacks depending on the class-wise vulnerability:

\begin{equation}
\label{eq:trix}
\min_{\theta} \; \mathbb{E}_{(\bm{x}, y) \sim \mathcal{D}} \left[
    \mathcal{L}_{\mathrm{CE}}(f_\theta(\bm{x}), y) +
    \beta \cdot 
    \begin{cases}
        \text{KL}(f_\theta(\bm{x}) \| f_\theta(\bm{x}^{\mathrm{adv}})), & \text{if } \pi(y) = \text{untargeted} \\
        \text{KL}(f_\theta(\bm{x}) \| f_\theta(\bm{x}_t^{\mathrm{adv}})), & \text{if } \pi(y) = \text{targeted} \\
    \end{cases}
\right]
\end{equation}

This unified objective enables the model to enforce stronger robustness for vulnerable (weak) classes through untargeted perturbations, while refining the decision boundaries of these high intrinsic diversity classes and promoting feature separability for strong classes with low intrinsic diversity through targeted perturbations in AT. The policy $\pi(y)$ can be instantiated using dynamic criteria such as inter-class feature similarity, margin collapse indicators, or empirical class-wise robustness measures.

\section{Related Work}
\paragraph{Fairness in Adversarial Training.}
Adversarial training (AT) is a widely adopted defense against adversarial examples, yet it often amplifies class-wise disparities in robustness, disproportionately harming underperforming or weak classes \citep{fairness_analysis, fairness_understanding, fairness_kdd, fairness_weighting}. To mitigate this, fairness-aware AT methods have been proposed, typically by reweighting class-wise loss or adversarial perturbations based on their class-wise robust performance. \textit{Fair Robust Learning (FRL)} dynamically adjusts loss weights and margins based on class-wise accuracy \citep{xu2021frl}. FAT addresses the robustness–fairness trade-off via regularizing variance in adversarial risk \citep{ma2021fat}. \textit{Balanced Adversarial Training (BAT)} \citep{sun2021bat} equalizes both source-class and target-class vulnerabilities. Entropy based approaches like \citep{wu2021entropy} enhance fairness using maximum entropy regularization, while \citep{wei2023cfa} and \citep{yue2023fairard} propose class-wise calibrated training and reweighting strategies based on vulnerability, respectively.

Theoretical analyses further support these trends. \citep{fairness_frl} attribute robust unfairness to class difficulty and propose scaled regularization. \citep{fairness_analysis} formalize the fairness–robustness trade-off, showing that disparity grows with attack strength (\(\epsilon\)). Several works draw analogies to long-tailed (LT) learning: \citep{fairness_weighting} and \citep{fairness_kdd} adapt LT balancing techniques to improve fairness in adversarial settings. Others such as \citep{fairness_understanding}, \citep{fairness_wat}, and \citep{fairness_bat} introduce adaptive strategies using validation feedback or fairness decomposition.

In contrast to prior class-centric approaches, we propose \textbf{TRIX}—a fairness-aware adversarial training framework that dynamically switches between targeted and untargeted adversaries based on class-wise feature similarity distance. TRIX further incorporates similarity driven loss weighting to prioritize the model optimization for vulnerable classes while increasing feature variability in strong classes. This design enables TRIX to better align training pressure with class difficulty, improving inter-class fairness and worst-case robust accuracy.

\section{Empirical Analysis}
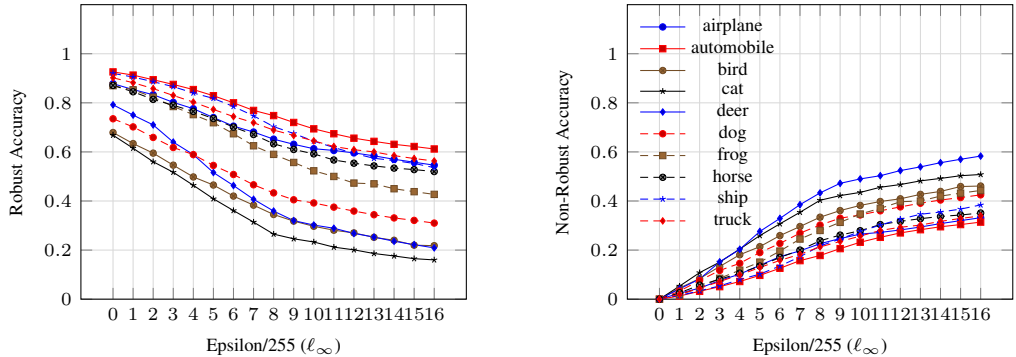
\begin{figure}[t]
\centering

\begin{tikzpicture}
\begin{axis}[
    hide axis,
    xmin=0, xmax=1,
    ymin=0, ymax=1,
    legend columns=5,
    legend style={
        draw=none,
        font=\scriptsize,
        column sep=0.5cm,
        anchor=north,
        at={(0.5, 1.0)}
    }
]
\end{axis}
\end{tikzpicture}

\vspace{-1em}  

\begin{minipage}[t]{0.48\textwidth}
\centering
\begin{tikzpicture}
\begin{axis}[
    width=\textwidth,
    height=5.5cm,
    xlabel={Epsilon/255 ($\ell_\infty$)},
    ylabel={Robust Accuracy},
    xtick={0,...,16},
     ytick={0.0, 0.2, 0.4, 0.6, 0.8,1.0},
    ymin=0.0,
    ymax=1.2,
    tick label style={font=\scriptsize},
    label style={font=\scriptsize},
    legend style={
        font=\scriptsize,
        at={(0.99,0.99)},
        anchor=north east,
        draw=none,
        fill=none,
        legend columns=1,
        row sep=-2pt
    },
    grid=both,
    grid style={gray!30},
    mark size=1.2pt,
    cycle list name=color
]
\addplot coordinates {(0,0.8780)(1,0.854) (2,0.833) (3,0.802) (4,0.776) (5,0.740) (6,0.705) (7,0.681) (8,0.652) (9,0.631) (10,0.613) (11,0.606) (12,0.596) (13,0.584) (14,0.570) (15,0.558) (16,0.546)};
\addplot coordinates {(0.0000,0.9260)(1,0.913) (2,0.894) (3,0.875) (4,0.854) (5,0.829) (6,0.800) (7,0.769) (8,0.748) (9,0.720) (10,0.694) (11,0.674) (12,0.656) (13,0.643) (14,0.631) (15,0.622) (16,0.612)};
\addplot coordinates {(0.0000,0.6790)(1,0.634) (2,0.595) (3,0.546) (4,0.498) (5,0.465) (6,0.420) (7,0.383) (8,0.345) (9,0.318) (10,0.297) (11,0.281) (12,0.270) (13,0.252) (14,0.240) (15,0.220) (16,0.218)};
\addplot coordinates {(0.0000,0.6680) (1,0.615) (2,0.560) (3,0.517) (4,0.464) (5,0.409) (6,0.361) (7,0.314) (8,0.265) (9,0.246) (10,0.233) (11,0.212) (12,0.201) (13,0.186) (14,0.176) (15,0.165) (16,0.160)};
\addplot coordinates {(0.0000,0.7920)(1,0.750) (2,0.710) (3,0.640) (4,0.589) (5,0.516) (6,0.463) (7,0.407) (8,0.359) (9,0.320) (10,0.302) (11,0.289) (12,0.268) (13,0.253) (14,0.236) (15,0.222) (16,0.209)};
\addplot coordinates {(0.0000,0.7350) (1,0.702) (2,0.659) (3,0.618) (4,0.589) (5,0.545) (6,0.508) (7,0.466) (8,0.433) (9,0.405) (10,0.392) (11,0.375) (12,0.359) (13,0.344) (14,0.331) (15,0.321) (16,0.310)};
\addplot coordinates {(0.0000,0.8700) (1,0.853) (2,0.823) (3,0.785) (4,0.752) (5,0.719) (6,0.674) (7,0.625) (8,0.590) (9,0.557) (10,0.523) (11,0.500) (12,0.473) (13,0.470) (14,0.450) (15,0.438) (16,0.427)};
\addplot coordinates {(0.0000,0.8710) (1,0.845) (2,0.814) (3,0.790) (4,0.765) (5,0.736) (6,0.701) (7,0.671) (8,0.633) (9,0.610) (10,0.592) (11,0.567) (12,0.554) (13,0.543) (14,0.534) (15,0.528) (16,0.520)};
\addplot coordinates {(0.0000,0.9200) (1,0.905) (2,0.886) (3,0.866) (4,0.840) (5,0.818) (6,0.785) (7,0.747) (8,0.703) (9,0.675) (10,0.644) (11,0.618) (12,0.594) (13,0.574) (14,0.565) (15,0.553) (16,0.536)};
\addplot coordinates {(0.0000,0.9020)(1,0.882) (2,0.858) (3,0.830) (4,0.803) (5,0.773) (6,0.744) (7,0.719) (8,0.689) (9,0.667) (10,0.645) (11,0.622) (12,0.609) (13,0.600) (14,0.585) (15,0.573) (16,0.563)};
\end{axis}
\end{tikzpicture}
\end{minipage}
\hfill
\begin{minipage}[t]{0.48\textwidth}
\centering
\begin{tikzpicture}
\begin{axis}[
    width=\textwidth,
    height=5.5cm,
    xlabel={Epsilon/255 ($\ell_\infty$)},
    ylabel={Non-Robust Accuracy},
    xtick={0,...,16},
    ytick={0.0, 0.2, 0.4, 0.6, 0.8,1.0},
    ymin=0.0,
    ymax=1.2,
    tick label style={font=\scriptsize},
    label style={font=\scriptsize},
    legend style={
        font=\scriptsize,
        at={(0.01,0.99)},
        anchor=north west,
        draw=none,
        fill=none,
        legend columns=1,
        row sep=-2pt
    },
    grid=both,
    grid style={gray!30},
    mark size=1.2pt,
    cycle list name=color
]
\addplot coordinates {(0.0000,0.0000)(1,0.024) (2,0.045) (3,0.076) (4,0.102) (5,0.138) (6,0.173) (7,0.197) (8,0.226) (9,0.247) (10,0.265) (11,0.272) (12,0.282) (13,0.294) (14,0.308) (15,0.320) (16,0.332)};
\addlegendentry{airplane}
\addplot coordinates {(0.0000,0.0000)(1,0.013) (2,0.032) (3,0.051) (4,0.072) (5,0.097) (6,0.126) (7,0.157) (8,0.178) (9,0.206) (10,0.232) (11,0.252) (12,0.270) (13,0.283) (14,0.295) (15,0.304) (16,0.314)};
\addlegendentry{automobile}
\addplot coordinates {(0.0000,0.0000)(1,0.045) (2,0.084) (3,0.133) (4,0.181) (5,0.214) (6,0.259) (7,0.296) (8,0.334) (9,0.361) (10,0.382) (11,0.398) (12,0.409) (13,0.427) (14,0.439) (15,0.459) (16,0.461)};
\addlegendentry{bird}
\addplot coordinates {(0.0000,0.0000)(1,0.053) (2,0.108) (3,0.151) (4,0.204) (5,0.259) (6,0.307) (7,0.354) (8,0.403) (9,0.422) (10,0.435) (11,0.456) (12,0.467) (13,0.482) (14,0.492) (15,0.503) (16,0.508)};
\addlegendentry{cat}
\addplot coordinates {(0.0000,0.0000)(1,0.042) (2,0.082) (3,0.152) (4,0.203) (5,0.276) (6,0.329) (7,0.385) (8,0.433) (9,0.472) (10,0.490) (11,0.503) (12,0.524) (13,0.539) (14,0.556) (15,0.570) (16,0.583)};
\addlegendentry{deer}
\addplot coordinates {(0.0000,0.0000)(1,0.033) (2,0.076) (3,0.117) (4,0.146) (5,0.190) (6,0.227) (7,0.269) (8,0.302) (9,0.330) (10,0.343) (11,0.360) (12,0.376) (13,0.391) (14,0.404) (15,0.414) (16,0.425)};
\addlegendentry{dog}
\addplot coordinates {(0.0000,0.0000)(1,0.017) (2,0.047) (3,0.085) (4,0.118) (5,0.151) (6,0.196) (7,0.245) (8,0.280) (9,0.313) (10,0.347) (11,0.370) (12,0.397) (13,0.400) (14,0.420) (15,0.432) (16,0.443)};
\addlegendentry{frog}
\addplot coordinates {(0.0000,0.0000)(1,0.026) (2,0.057) (3,0.081) (4,0.106) (5,0.135) (6,0.170) (7,0.200) (8,0.238) (9,0.261) (10,0.279) (11,0.304) (12,0.317) (13,0.328) (14,0.337) (15,0.343) (16,0.351)};
\addlegendentry{horse}
\addplot coordinates {(0.0000,0.0000)(1,0.015) (2,0.034) (3,0.054) (4,0.080) (5,0.102) (6,0.135) (7,0.173) (8,0.217) (9,0.245) (10,0.276) (11,0.302) (12,0.326) (13,0.346) (14,0.355) (15,0.367) (16,0.384)};
\addlegendentry{ship}
\addplot coordinates {(0.0000,0.0000)(1,0.020) (2,0.044) (3,0.072) (4,0.099) (5,0.129) (6,0.158) (7,0.183) (8,0.213) (9,0.235) (10,0.257) (11,0.280) (12,0.293) (13,0.302) (14,0.317) (15,0.329) (16,0.339)};
\addlegendentry{truck}
\end{axis}
\end{tikzpicture}
\end{minipage}

\caption{
Class-wise performance under PGD attack across increasing $\ell_\infty$ perturbation strengths. 
\textbf{Left:} Robust accuracy on adversarial examples. 
\textbf{Right:} Non-robust feature reliance (accuracy drop with respect to the accuracy on clean data).
}
\label{fig:classwise_pgd}
\end{figure}

In this section, we present the motivation for integrating both targeted and untargeted adversarial perturbations within adversarial training. We begin by analyzing how the strength of adversarial attacks differentially impacts distinct class groups, focusing on the suppression of non-robust features under varying perturbation margins. Next, we examine class-wise performance under purely targeted versus untargeted training objectives to examine their individual capabilities. 
\vspace{-0.2cm}
\paragraph{Impact of Attack Strength.} As illustrated in Fig.~\ref{fig:classwise_pgd}, increasing the perturbation strength of PGD attacks measured with in $\ell_\infty$ perturbation bound $\epsilon$ reveals asymmetric vulnerabilities across classes when models are trained with conventional TRADES. While overall robust accuracy consistently declines as $\epsilon$ increases, the rate of degradation varies substantially by class. For instance, \textit{automobile} and \textit{truck} classes exhibit a relatively gradual decline, maintaining notable levels of robustness under stronger perturbations. In contrast, classes with high intra-class variability such as \textit{cat}, \textit{bird}, and \textit{deer} suffer rapid drops in robust accuracy, underscoring their inherent susceptibility.

Interestingly, non-robust feature accuracy measured by drop in accuracy from clean samples compared to robust samples intensifies with increasing $\epsilon$, particularly among these more vulnerable classes. This suggests that such classes are less capable of suppressing non-robust features when exposed to fixed-strength adversarial perturbations, leading to disproportionately severe degradation in robust performance.

This observed imbalance exposes a core limitation of standard adversarial training: uniform perturbation budgets can inadvertently amplify robustness disparities across classes. Recent methods such as FAIR-TAT~\citep{targeted_adv_training}, CFA~\citep{wei2023cfa}, and DAFA~\citep{lee2024dafa} have addressed this by adapting either the strength or target of adversarial perturbations based on class-specific traits. Motivated by these insights, our approach introduces a novel framework that dynamically adjusts both the \emph{type} and \emph{magnitude} of adversarial examples during training. Specifically, we leverage inter-class feature similarity to guide this adaptation, facilitating more balanced suppression of non-robust features and fostering equitable robustness across the full spectrum of classes.

\usetikzlibrary{patterns}
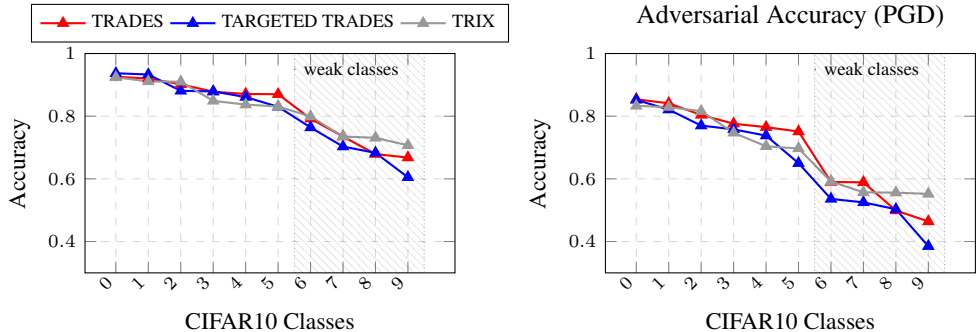
\begin{figure}
\centering
\begin{tikzpicture}
\begin{groupplot}[
    group style={
        group size=2 by 1,
        horizontal sep=2cm,
    },
    width=6.5cm,
    height=4.5cm,
    ymin=0.3, ymax=1.0,
    xlabel={CIFAR10 Classes},
    ylabel={Accuracy},
    xtick={0,...,9},
    xticklabel style={rotate=45, anchor=east, font=\scriptsize},
    tick label style={font=\scriptsize},
    title style={font=\normalsize},
    label style={font=\small},
    legend style={at={(0.5,1.05)}, anchor=south, legend columns=3, font=\scriptsize},
    grid=both,
    grid style={dashed,gray!30},
    cycle list name=color list,
]

\nextgroupplot[title={Clean Accuracy}]
\legend{TRADES, TARGETED TRADES, TRIX}
\addplot+[mark=triangle*, thick] coordinates {
    (0,0.926)  
    (1,0.920)  
    (2,0.902)  
    (3,0.878)  
    (4,0.871)  
    (5,0.870)  
    (6,0.792)  
    (7,0.735)  
    (8,0.679)  
    (9,0.668)  
};

\addplot+[mark=triangle*, thick] coordinates {
    (0,0.937)  
    (1,0.933)  
    (2,0.881)  
    (3,0.879)  
    (4,0.861)  
    (5,0.830)  
    (6,0.764)  
    (7,0.703)  
    (8,0.683)  
    (9,0.605)  
};

\addplot+[gray!80, mark=triangle*, thick] coordinates {
    (0,0.924)  
    (1,0.911)  
    (2,0.910)  
    (3,0.849)  
    (4,0.837)  
    (5,0.831)  
    (6,0.799)  
    (7,0.735)  
    (8,0.731)  
    (9,0.707)  
};
\addplot [pattern=north west lines,densely dotted,opacity=0.3] coordinates {
         (5.5,\pgfkeysvalueof{/pgfplots/ymax})
         (9.5,\pgfkeysvalueof{/pgfplots/ymax})
        }
        \closedcycle;
\draw [] (axis cs:5.5, 0.95)-- +(0pt,0pt) node[right] {\scriptsize weak classes};
\nextgroupplot[title={Adversarial Accuracy (PGD)}]

\addplot+[mark=triangle*, thick] coordinates {
    (0,0.854)  
    (1,0.841)  
    (2,0.804)  
    (3,0.776)  
    (4,0.765)  
    (5,0.751)  
    (6,0.590)  
    (7,0.589)  
    (8,0.499)  
    (9,0.464)  
};

\addplot+[mark=triangle*, thick] coordinates {
    (0,0.852)  
    (1,0.821)  
    (2,0.770)  
    (3,0.758)  
    (4,0.738)  
    (5,0.650)  
    (6,0.536)  
    (7,0.525)  
    (8,0.503)  
    (9,0.385)  
};

\addplot+[gray!80, mark=triangle*, thick] coordinates {
    (0,0.833)  
    (1,0.829)  
    (2,0.816)  
    (3,0.747)  
    (4,0.704)  
    (5,0.697)  
    (6,0.592)  
    (7,0.557)  
    (8,0.556)  
    (9,0.552)  
};
\addplot [pattern=north west lines,densely dotted,opacity=0.3] coordinates {
         (5.5,\pgfkeysvalueof{/pgfplots/ymax})
         (9.5,\pgfkeysvalueof{/pgfplots/ymax})
        }
        \closedcycle;
\draw [] (axis cs:5.5, 0.95)-- +(0pt,0pt) node[right] {\scriptsize weak classes};

\end{groupplot}
\end{tikzpicture}
\caption{Comparison of class-wise clean and adversarial (PGD) accuracy for TRADES and TARGETED TRADES on CIFAR-10. Classes are sorted according to their clean TRADES accuracy in descending order. \textbf{Left}: Clean accuracy. \textbf{Right}: PGD adversarial accuracy with $\epsilon = 4/255$.
Performance on weak classes is substantially improved with TRIX.
} 
\label{fig:classwise_targeted_untargeted_comparison}
\end{figure}

\paragraph{Targeted vs. Untargeted Adversarial Training.} Building on the observation that different classes exhibit varying levels of vulnerability to adversarial attacks, we explore the distinct effects of \textbf{targeted} and \textbf{untargeted} adversarial training. We refer to \textbf{TRADES} as the standard untargeted adversarial training method, and introduce \textbf{TARGETED TRADES} as its targeted counterpart, which applies the TRADES objective using targeted perturbations.

As illustrated in Figure~\ref{fig:classwise_targeted_untargeted_comparison}, the \textit{clean accuracy} plot (left) shows that both TRADES and TARGETED TRADES achieve similar overall performance. However, they differ significantly at the class level. TRADES demonstrates a relative advantage on \textit{weaker or more ambiguous classes} such as \textit{cat} and \textit{dog}, maintaining higher clean accuracy. In contrast, TARGETED TRADES underperforms on these vulnerable categories, likely because targeted attacks are weaker and also exploit semantic similarities to mislead predictions, which is particularly harmful for such classes. 

Conversely, TARGETED TRADES shows marginal improvements or maintains performance on already \textit{strong classes} such as \textit{automobile}, \textit{airplane}, and \textit{truck}. 
This trend persists in the \textit{adversarial accuracy} plot (right), where the targeted variant continues to lag behind TRADES on vulnerable classes but shows slight gains on robust ones. This is due increase in feature separability by sampling the targets uniformly.

These observations indicate that \textbf{untargeted training encourages broader robustness}, especially for confusing or overlapping classes, while \textbf{targeted training promotes localized robustness} around dominant or high-performing categories. In applications where robustness on critical classes is a priority, targeted training may offer practical benefits. However, for \textbf{balanced performance or fairness across all classes}, a hybrid strategy that integrates both targeted and untargeted adversaries during training could be more effective, which is shown by TRIX in Figure~\ref{fig:classwise_targeted_untargeted_comparison}.
\section{Adversarially Fair TRADES via Mixed Adversarial Training} \label{methodlogy}

We propose \textbf{TRIX} (\textit{\textbf{Tr}ading Adversarial Fairness via M\textbf{ix}ed Adversarial Training}), a novel adversarial training framework that enhances both robustness and fairness by applying feature-aware, class-specific perturbations. 
Our approach consists of three core components: 
(1) computation of adaptive class weights using instance-level feature similarity distance; 
(2) mixed adversarial training via class-dependent perturbation objectives; and 
(3) class-aware loss weighting and perturbation scaling to emphasize underperforming classes.

\paragraph{Adaptive Class Weights via Feature Similarity.} To guide adversarial training towards vulnerable classes, we compute adaptive class weights based on inter-class similarity derived from predicted probabilities on clean samples $x_i$ for each instance. Given $N$ samples with predicted probability vectors $\mathbf{p}_i \in \mathbb{R}^C$ and labels $y_i \in \{0, \ldots, C{-}1\}$, we first compute the average predicted distribution for each class $c$:
\begin{equation}
\bar{\mathbf{p}}_c = \frac{1}{|\mathcal{D}_c|} \sum_{i \in \mathcal{D}_c} \mathbf{p}_i,
\quad \text{where } \mathcal{D}_c = \{i \mid y_i = c\}.   
\end{equation}

Stacking the class-wise average predictions $\bar{\mathbf{p}}_c$ row-wise forms the class similarity matrix $\mathbf{S} \in \mathbb{R}^{C \times C}$, where $\mathbf{S}_{c,j} = \bar{\mathbf{p}}_c[j]$, $S_{c,j}$ indicates the average likelihood of samples from class $c$ being predicted as class $j$. The diagonal entry $S_{c,c}$ captures the model’s self-confidence on class $c$, while off-diagonal entries reflect inter-class similarity distances. To ensure numerical stability, we regularize the diagonal using $\mu=1e^-8$:
\begin{equation}
  \mathbf{S} \leftarrow \mathbf{S} + \mu \cdot \mathbf{I}.  
\end{equation}

We then define a class weight vector $\mathbf{w} \in \mathbb{R}^C$ that emphasizes classes with low similarity distances, i.e., high confusion and low confidence:
\begin{equation}
w_c = 1 + \lambda \sum_{j \neq c}
\begin{cases}
S_{c,j} \cdot S_{j,j}, & \text{if } S_{c,c} < S_{j,j}, \\
- S_{j,c} \cdot S_{c,c}, & \text{otherwise},
\end{cases}
\label{eq:weight_update}
\end{equation}

where $\lambda$ controls the strength of weight adjustment. Intuitively, classes with weaker self-confidence and greater overlap with other classes are upweighted, while dominant classes are softly penalized, promoting fairness in robustness distribution.

\paragraph{Mixed TRADES Loss with Adaptive Targeting.} To incorporate class-aware perturbations into adversarial training, We implement a mixed TRADES loss, where the choice between targeted and untargeted adversarial objectives is made per sample based on evaluated adaptive class weights. For each training instance $\bm{x}_i$ with label $y_i$, we define a binary indicator $\mathbbm{1}_{\text{targeted}}^{(i)} \in \{0, 1\}$ as:
\begin{equation}
\mathbbm{1}_{\text{targeted}}^{(i)} =
\begin{cases}
1, & \text{if } w_{y_i} \leq \textit{avg}(\mathbf{w}), \\
0, & \text{otherwise},
\end{cases}
\end{equation}
where $\textit{avg}(\mathbf{w})$ is the average of the class weight vector $\mathbf{w}$ over the batch of samples. This binary indicator ensures targeted attacks are assigned to confident (low-weight) classes, and untargeted attacks to underperforming (high-weight) ones.

Let $\bm{x}_i^{\text{adv}}$ denote the untargeted adversarial example generated for $\bm{x}_i$, and let $\bm{x}_{ti}^{\text{adv}}$ be the targeted adversarial example attacked toward a randomly chosen target label $\tilde{y}_i \neq y_i$. The mixed adversarial loss is defined as:
\begin{equation}
\mathcal{L}_{\text{adv}} = \frac{1}{B} \sum_{i=1}^{B} \left[
    \mathbbm{1}_{\text{targeted}}^{(i)} \cdot \mathrm{KL}\left(f_\theta(\bm{x}_{ti}^{\text{adv}}) \parallel f_\theta(\bm{x}_i)\right) +
    \left(1 - \mathbbm{1}_{\text{targeted}}^{(i)}\right) \cdot \mathrm{KL}\left(f_\theta(\bm{x}_i) \parallel f_\theta(\bm{x}_i^{\text{adv}})\right)
\right],
\label{eq:mixed_adv_loss}
\end{equation}

where $\tilde{y}_i \ne y_i$ is a randomly sampled target class for the targeted objective, and $f_\theta(\cdot)$ denotes the model’s predicted logits. $B$ denotes the batch of samples. For targeted and untargeted perturbation samples, we compute the KL divergence to enforce consistency between clean and adversarial predictions. Furthermore, we scale the TRADES perturbation radius per class sample using the corresponding class weight $w_{y_i}$, allowing weaker classes to explore a wider adversarial neighborhood while constraining stronger ones. This encourages a more balanced and class-aware robustness across the model.

\paragraph{Final Training Objective.} The complete TRIX objective combines the natural loss with the mixed adversarial loss, weighted by the adaptive class weights:

\begin{equation}
\mathcal{L}_{\text{total}} = \mathbb{E}_{i} \left[ w_{y_i} \cdot \mathrm{CE}(f_\theta(\bm{x}_i), y_i) \right]
+ \beta \cdot \mathcal{L}_{\text{adv}},
\label{eq:final_loss}
\end{equation}

where $\beta$ controls the trade-off between standard accuracy and adversarial robustness. The complete algorithm of TRIX can be found in Appendix \ref{algoritm}.

\begin{table}[t]
\centering
\small
\setlength{\tabcolsep}{5pt}
\caption{AutoAttack evaluation on ResNet-18 for different robust fairness approaches with $\epsilon=8/255$. }
\begin{subtable}{\textwidth}
\centering
\caption{CIFAR-10}
\resizebox{\textwidth}{!}{
\begin{tabular}{l|ccc|ccc}
\toprule
\textbf{Method} & \textbf{Avg Clean} & \textbf{Worst Clean} & $\varrho_{\text{clean}}$ & \textbf{Avg Robust} & \textbf{Worst Robust} & $\varrho_{\text{robust}}$ \\
\midrule
TRADES              & 82.17 ± 0.12 & 66.21 ± 1.21 & 0.00 & 49.74 ± 0.20 & 23.11 ± 0.80 & 0.00 \\
TRADES + FRL        & 82.89 ± 0.35 & 71.86 ± 0.92 & \textbf{0.07} & 45.32 ± 0.26 & 26.71 ± 0.83 & 0.06 \\
TRADES + BAT        & \textbf{86.91 ± 0.09} & \textbf{73.42 ± 1.28} & 0.05 & 46.01 ± 0.19 & 18.50 ± 1.35 & -0.12 \\
TRADES + WAT        & 79.97 ± 0.32 & 70.21 ± 1.63 & 0.03 & 46.16 ± 0.23 & 28.81 ± 1.41 & 0.17 \\
TRADES + CFA        & 79.65 ± 0.14 & 64.98 ± 0.21 & -0.01 & \textbf{50.81 ± 0.17} & 26.89 ± 0.29 & 0.14 \\
TRADES + FAIRTAT    & 79.45 ± 0.11 & 67.98 ± 0.34 & -0.06 & 50.32 ± 0.22 & 28.13 ± 0.28 & 0.20\\
TRADES + DAFA       & 81.90 ± 0.29 & 67.90 ± 1.21 &  0.02 & 49.45 ± 0.22 & 30.20 ± 1.13 & 0.30 \\
TRADES + TRIX       & 82.68 ± 0.23 & 69.98 ± 1.06 & 0.05 & 49.09 ± 0.24 & \textbf{32.65 ± 1.07} & \textbf{0.39} \\
\bottomrule
\end{tabular}
\label{tab:cifar10}
}
\end{subtable}

\vspace{1em}

\begin{subtable}{\textwidth}
\centering
\caption{CIFAR-100}
\resizebox{\textwidth}{!}{
\begin{tabular}{l|ccc|ccc}
\toprule
\textbf{Method} & \textbf{Avg Clean} & \textbf{Worst Clean} & $\varrho_{\text{clean}}$ & \textbf{Avg Robust} & \textbf{Worst Robust} & $\varrho_{\text{robust}}$ \\
\midrule
TRADES            & 58.13 ± 0.20 & 17.21 ± 1.12 & 0.00 & 25.63 ± 0.15 & 1.49 ± 0.91 & 0.00\\
TRADES + FRL      & 57.81 ± 0.21 & 18.86 ± 1.90 & 0.09 & 25.09 ± 0.27 & 1.81 ± 0.63 & 0.19\\
TRADES + BAT      & \textbf{61.98 ± 0.19} & \textbf{22.61 ± 1.97} & \textbf{0.24} & 23.41 ± 0.20 & 0.38 ± 0.61 & -0.65 \\
TRADES + WAT      & 53.56 ± 0.37 & 17.14 ± 2.01 & -0.07  & 22.31 ± 0.26 & 1.96 ± 0.12 & 0.18 \\
TRADES + CFA      & 57.63 ± 1.20 & 15.22 ± 1.39 & -0.10 & 23.49 ± 0.27 & 1.71 ± 1.66 & 0.06 \\
TRADES + FAIRTAT  & 56.47 ± 1.32 & 15.98 ± 1.08 & -0.04 & 23.65 ± 0.19 & 1.74 ± 1.91 & 0.09 \\
TRADES + DAFA     & 57.97 ± 0.10 & 18.61 ± 0.36 & 0.07 & 25.11 ± 0.17 & 2.41 ± 0.45 & 0.59\\
TRADES + TRIX     & 58.90 ± 0.17 & 19.89 ± 0.28 & 0.14 & \textbf{25.38 ± 0.16} & \textbf{2.89 ± 0.46} & \textbf{0.92}\\
\bottomrule
\end{tabular}
\label{tab:cifar100}
}
\end{subtable}

\vspace{1em}

\begin{subtable}{\textwidth}
\centering
\caption{STL-10}
\resizebox{\textwidth}{!}{
\begin{tabular}{l|ccc|ccc}
\toprule
\textbf{Method} & \textbf{Avg Clean} & \textbf{Worst Clean} & $\varrho_{\text{clean}}$ & \textbf{Avg Robust} & \textbf{Worst Robust} & $\varrho_{\text{robust}}$ \\
\midrule
TRADES  & 61.32 ± 0.67 & 38.12 ± 1.86 & 0.00 &  29.93 ± 0.63 & 7.61 ± 1.21 & 0.00 \\
TRADES + FRL      & 57.41 ± 0.20 & 30.98 ± 1.67 & -0.12   & 28.67 ± 0.38 & 7.91 ± 0.38 & 0.00 \\
TRADES + BAT      & 59.41 ± 1.07 & 35.98 ± 2.52 &  -0.02 & 24.15 ± 1.03 & 3.67 ± 0.93 & -0.32 \\
TRADES + WAT      & 53.44 ± 1.70 & 31.11 ± 1.39 & -0.05 & 26.76 ± 0.31 & 6.78 ± 0.71 & 0.00 \\
TRADES + CFA      & 60.67 ± 0.59 & 38.43 ± 0.94 & 0.00 & \textbf{31.91 ± 0.20} & 7.62 ± 0.32 & -0.06 \\
TRADES + DAFA     & 60.23 ± 0.61 & 41.96 ± 2.32 & 0.08 & 29.87 ± 0.51 & 10.57 ± 1.28 & 0.38\\
TRADES + TRIX     & \textbf{61.52 ± 0.63} & \textbf{43.18 ± 2.31} & \textbf{0.12} & 29.89 ± 0.39 & \textbf{10.97 ± 1.41} & \textbf{0.44} \\
\bottomrule
\end{tabular}
\label{tab:stl10}
}
\end{subtable}
\label{tab:evaluations}
\end{table}

\section{Experiments} \label{exps}

\paragraph{Datasets and Baseline Details.}\label{exp:paragraph_implementation_details}
To demonstrate the effectiveness of TRIX framework, we conduct experiments on standard image classification benchmarks: CIFAR-10 \citep{cifar10}, CIFAR-100, and STL-10 \citep{coates2011analysis}. We utilize TRADES as the baseline adversarial training algorithm and adopt ResNet-18 \citep{he_deep_2016} as the model architecture. For comparison, we include state-of-the-art adversarial training frameworks that focus on adversarial fairness, including recent methods such as DAFA \citep{lee2024dafa}, FAIR-TAT \citep{medi2025fair}, and CFA \citep{wei2023cfa}, as well as earlier approaches like BAT \citep{fairness_bat}, WAT \citep{li2023wat}, and FRL \citep{fairness_frl}. All comparison methods are built upon the TRADES framework to ensure fair and consistent evaluation. Additional details and descriptions of the comparison methods are provided in the Appendix \ref{additional_details}.
\vspace{-0.2cm}
\paragraph{Hyperparameters, Implementation Details and Metrics.}\label{hyperparameters}
Following prior work~\citep{lee2024dafa}, we apply perturbation margin adjustments and loss weighting after a warm-up phase of $\tau = 70$ epochs. The weight strength adjustment coefficient $\lambda$ is set to 1.0 for CIFAR-10 and 1.5 for CIFAR-100 and STL-10. Robustness and fairness are evaluated using adaptive AutoAttack~\citep{attack_aaa}. Due to the higher variance in class-wise accuracy, each experiment is repeated with 5 different seeds, reporting the mean accuracy over the last 5 epochs per run (averaged across 25 models). For fair comparison, we use official implementations of all baselines. TRIX is trained with SGD (momentum 0.9, weight decay $5 \times 10^{-4}$, Nesterov enabled), using an initial learning rate of 0.1 with step decay (reduced by 10× at epochs $T{-}10$ and $T{-}5$ for $T=110$ total epochs). Training is conducted on RTX A6000 GPUs. We assess adversarial fairness using the \textit{minimum class-wise accuracy} on both clean and robust samples, capturing worst-case class performance. To ensure overall performance is maintained, we also report the \textit{average accuracy} in both settings. Additionally, we include the metric $\varrho$, which measures the \textit{relative change in worst class accuracy} with baseline TRADES compared to the \textit{relative change in overall accuracy} with baseline TRADES. A higher $\varrho$ indicates improved fairness preservation relative to overall performance. Additional implementation details and details on $\varrho$ in Appendix \ref{additional_details}.

\paragraph{Adversarial Fairness Performance.}
Table~\ref{tab:evaluations} presents AutoAttack evaluations of ResNet-18 models trained with different adversarial fairness approaches on CIFAR-10, CIFAR-100, and STL-10 using perturbation $\epsilon = 8/255$ threat model. We compare TRADES with fairness-oriented baselines: DAFA~\citep{lee2024dafa}, FRL~\citep{fairness_frl}, BAT~\citep{fairness_bat}, WAT~\citep{fairness_wat}, CFA~\citep{wei2023cfa}, and FAIR-TAT~\citep{medi2025fair}. We report both average and worst-class accuracies on clean and adversarial samples to assess fairness-robustness trade-offs. On CIFAR-10 (Table~\ref{tab:cifar10}), most methods improve worst-class clean accuracy over TRADES. BAT achieves the highest average clean accuracy (86.91\%) and worst clean accuracy (73.42\%), but performs poorly on adversarial examples. FRL similarly improves fairness on clean data but at the cost of overall robust accuracy. TRIX stands out by improving both clean worst-class accuracy (69.98\%) and worst robust accuracy (32.65\%) while maintaining strong average performance indicated by high value of $\varrho_{clean}$, offering a better balance between robustness and fairness. 

Tables~\ref{tab:cifar100} and~\ref{tab:stl10} report evaluations on CIFAR-100 and STL-10, which include more classes and higher resolution, respectively. On CIFAR-100, TRIX achieves the best robust worst-class accuracy (2.89\%), outperforming all baselines while maintaining comparable average performance. On STL-10, TRIX again obtains the highest robust worst-class accuracy (10.97\%) while preserving clean and robust accuracy close to the TRADES baseline. TRIX also acheieves better fairness on clean samples for CIFAR-100 (19.89\%) and STL-10 (43.18\%). For both these datasets, TRIX also achieves higher values of $\varrho_{clean}$ and $\varrho_{robust}$. These results highlight the effectiveness of TRIX in enhancing adversarial fairness across diverse datasets without compromising robustness. Additional experiments and comparisons with approach are found in Appendix \ref{different_arch}.
\vspace{-0.2cm}
\paragraph{Design Choices of TRIX.}
\begin{wraptable}{r}{0.6\linewidth}
\vspace{-1em}
\caption{Ablation and Comparison to DAFA.}
\label{tab:ablation}
\centering
\begin{tabular}{lcccc}
\toprule
\textbf{Method} & \textbf{Avg} & \textbf{Worst} & \textbf{Avg} & \textbf{Worst} \\
                & \textbf{Clean} & \textbf{Clean} & \textbf{Robust} & \textbf{Robust} \\
\midrule
DAFA-Basic          &         82.41 & 66.80 & 54.46 & 26.80 \\
TRIX-Uniform    &         82.11 & 66.60 & 53.45 & 30.70 \\
TRIX-Basic      & 82.43 & 66.70 & 53.86 & 28.23\\
DAFA-Uniform    &         81.85 & 65.70 & \textbf{54.56} & 31.20 \\
DAFA            &         82.17 & 67.50 & 54.16 & 36.50 \\
TRIX            & \textbf{82.58} & \textbf{68.90} & 53.21 & \textbf{38.90}\\
\bottomrule
\end{tabular}
\vspace{-1em}
\end{wraptable}
To evaluate the contribution of each component in TRIX, we perform an ablation study comparing the following variants: \textbf{TRIX-Basic}, which removes perturbation margin and loss weight adjustments, and \textbf{TRIX-Uniform}, which applies uniform loss weights. These are compared against the standard TRADES baseline (\textbf{DAFA-Basic}) and DAFA~\citep{lee2024dafa}, which incorporates similar training modifications in its respective variants. Table~\ref{tab:ablation} reports PGD evaluations on CIFAR-10 with $\epsilon=8/255$. Results indicate that each TRIX component contributes to improving the worst-class robust accuracy, with TRIX achieving the best fairness–robustness trade-off. More ablations on TRIX are found in Appendix \ref{ablation}.

\section{Conclusion} \label{conclusion}
In this work, We introduced \textbf{TRIX}, a novel adversarial training framework that dynamically combines targeted and untargeted adversaries to improve class-wise fairness in adversarial robustness. By adapting the type and strength of perturbations based on feature similarity and class vulnerability, TRIX enhances robustness for weak classes while preserving (or improving) performance on strong ones. It further incorporates adaptive loss weighting and perturbation scaling to align training pressure with class difficulty. Evaluations on CIFAR-10, CIFAR-100, and STL-10 demonstrate significant improvements in worst-class accuracy under both clean and robust settings, while maintaining the overall performance. Beyond empirical gains, TRIX contributes toward the development of reliable and fair models,critical for safety sensitive applications by mitigating bias and improving robustness under diverse conditions.

\paragraph{Limitations.}
TRIX improves fairness by balancing clean and robust accuracy across classes but does not significantly enhance overall robustness. It relies on feature similarity distances to identify weak and strong classes, though more sample-efficient alternatives could be explored.

\clearpage
\clearpage

\appendix
\section{Appendix}
\subsection{Algorithm of TRIX} \label{algoritm}
\begin{algorithm}[H]
\caption{TRIX Training Algorithm}
\label{alg:trix}
\Input{Dataset $\mathcal{D}$, Model $f_{\theta}$, Batch size $B$, Warm-up threshold $\tau$, Class-wise weight vector $\mathbf{w}$, Total epochs $T$}
\BlankLine

\textbf{Main TRIX Training Phase} \;
\For{$t \gets 1$ \KwTo $\tau$}{
  Sample batch  $B = \{(x_i, y_i)\}_{i=1}^B \sim \mathcal{D}$ \;
  \For{$i \gets 1$ \KwTo $B$}{
    Update class-wise clean prediction probabilities: $\bar{\mathbf{p}}_c \gets f(x_i, y_i)$ \;
  }
  Compute class-wise weight vector $\mathbf{w}$ \tcp*{Refer to Equation~\eqref{eq:weight_update}}

  \For{$i \gets 1$ \KwTo $B$}{
    \eIf{$w_{y_i} \leq \text{avg}(\mathbf{w})$}{
      Generate \textbf{targeted} adversarial examples $x_i^{adv}$ with target $\tilde{y} \neq y_i$ \;
    }{
      Generate \textbf{untargeted} adversarial examples $x_i^{adv}$ \;
    }
  }
  Update model parameters: $\theta \gets \arg\min_{\theta} \mathcal{L}_{\text{TRADES}}(x_i, x_i^{adv}, y_i)$ \;
}
Update class-wise robust prediction probabilities: $\bar{\mathbf{p}}_c \gets f(x_i^{adv}, y_i)$ \;
Recompute class-wise weights $\mathbf{w}$ using robust prediction probabilities\;
\BlankLine

\textbf{Warm-up Phase (TRADES)} \;
\For{$t \gets \tau + 1$ \KwTo $T$}{
  Sample batch $B = \{(x_i, y_i)\}_{i=1}^B \sim \mathcal{D}$ \;
  Generate $x_i^{adv}$ as in Main TRIX Training \;
  Update $\theta \gets \arg\min_{\theta} \mathcal{L}_{\text{TRADES}}(x_i, x_i^{adv}, y_i; w_{y_i})$ \;
}
\Return Trained model $f(\cdot; \theta)$ \;
\end{algorithm}

\paragraph{TRIX Algorithmic Overview.} 
TRIX is a mixed adversarial training method that assigns either targeted or untargeted adversarial examples to each class based on its vulnerability. The training scheme consists of two main phases. 

In the \textbf{Main TRIX Phase} (epochs $1$ to $\tau$): For each batch, the model computes a class-wise weight vector $\mathbf{w}$ using the feature prediction probabilities of clean samples. Classes with low weight (i.e., strong classes) are assigned \textit{targeted} adversarial examples, while classes with high weight (i.e., vulnerable/weak classes) receive \textit{untargeted} adversarial examples in AT. Refer Equation~\ref{eq:weight_update} for the information on class-wise weight vector calculation from feature prediction probabilities.

In the \textbf{Warm-up Phase} (epochs $\tau+1$ to $T$): The class-wise weight vector $\mathbf{w}$ that is computed at $\tau$ using prediction probabilities of adversarial (robust) examples, similar to DAFA~\citep{lee2024dafa} are utilized. These weights are used to scale both the loss and the adversarial perturbation strength, helping to further stabilize training and improve robustness. We only compute the adversarial class-wise weights to scale the loss and perturbation at warm-up interval $\tau=70$. The information on class-wise weights $\mathbf{w}$ and also optimization loss used in TRIX is discussed in Sec~\ref{methodlogy} of main paper.

Figure~\ref{fig:classwise_targeted_untargeted_comparison_unsorted} presents a comparison between the proposed \textsc{TRIX} approach, which integrates both targeted and untargeted adversarial examples during adversarial training and baseline models trained exclusively on either untargeted adversaries (\textsc{TRADES}) or targeted adversaries (\textsc{TARGETED TRADES}), using a ResNet-18 architecture. Notably, \textsc{TRIX} exhibits reduced performance disparity across classes, highlighting the effectiveness of strategically combining different types of adversaries in adversarial training.

\usetikzlibrary{patterns}
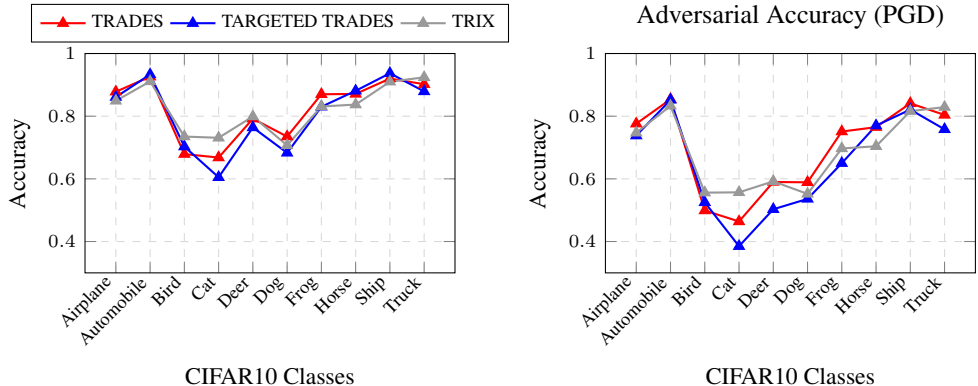
\begin{figure}
\centering
\begin{tikzpicture}
\begin{groupplot}[
    group style={
        group size=2 by 1,
        horizontal sep=2cm,
    },
    width=6.5cm,
    height=4.5cm,
    ymin=0.3, ymax=1.0,
    xlabel={CIFAR10 Classes},
    ylabel={Accuracy},
    xtick={0,...,9},
    xticklabels={Airplane,Automobile, Bird, Cat, Deer, Dog, Frog, Horse,Ship, Truck},
    xticklabel style={rotate=45, anchor=east, font=\scriptsize},
    tick label style={font=\scriptsize},
    title style={font=\normalsize},
    label style={font=\small},
    legend style={at={(0.5,1.05)}, anchor=south, legend columns=3, font=\scriptsize},
    grid=both,
    grid style={dashed,gray!30},
    cycle list name=color list,
]

\nextgroupplot[title={Clean Accuracy}]
\legend{TRADES, TARGETED TRADES, TRIX}
\addplot+[mark=triangle*, thick] coordinates {
    (0,0.878)  
    (1,0.926)  
    (2,0.679)  
    (3,0.668)  
    (4,0.792)  
    (5,0.735)  
    (6,0.870)  
    (7,0.871)  
    (8,0.920)  
    (9,0.902)  
};
\addplot+[mark=triangle*, thick] coordinates {
    (0,0.861)  
    (1,0.933)  
    (2,0.703)  
    (3,0.605)  
    (4,0.764)  
    (5,0.683)  
    (6,0.830)  
    (7,0.881)  
    (8,0.937)  
    (9,0.879)  
};
\addplot+[gray!80, mark=triangle*, thick] coordinates {
    (0,0.849)  
    (1,0.911)  
    (2,0.735)  
    (3,0.731)  
    (4,0.799)  
    (5,0.707)  
    (6,0.831)  
    (7,0.837)  
    (8,0.910)  
    (9,0.924)  
};
\nextgroupplot[title={Adversarial Accuracy (PGD)}]

\addplot+[mark=triangle*, thick] coordinates {
    (0,0.776)  
    (1,0.854)  
    (2,0.499)  
    (3,0.464)  
    (4,0.590)  
    (5,0.589)  
    (6,0.751)  
    (7,0.765)  
    (8,0.841)  
    (9,0.804)  
};
\addplot+[mark=triangle*, thick] coordinates {
    (0,0.738)  
    (1,0.852)  
    (2,0.525)  
    (3,0.385)  
    (4,0.503)  
    (5,0.536)  
    (6,0.650)  
    (7,0.770)  
    (8,0.821)  
    (9,0.758)  
};
\addplot+[gray!80, mark=triangle*, thick] coordinates {
    (0,0.747)  
    (1,0.833)  
    (2,0.556)  
    (3,0.557)  
    (4,0.592)  
    (5,0.552)  
    (6,0.697)  
    (7,0.704)  
    (8,0.816)  
    (9,0.829)  
};


\end{groupplot}
\end{tikzpicture}
\caption{Comparison of class-wise clean and adversarial (PGD) accuracy for TRADES and TARGETED TRADES on CIFAR-10. \textbf{Left}: Clean accuracy. \textbf{Right}: PGD adversarial accuracy with $\epsilon = 4/255$.
Performance on weak classes is substantially improved with TRIX.
    } 
\label{fig:classwise_targeted_untargeted_comparison_unsorted}
\end{figure}

\subsection{Additional Implementation and Experimental Details}\label{additional_details}

We evaluate our proposed TRIX framework on three standard image classification datasets: CIFAR-10, CIFAR-100~\citep{krizhevsky2009learning}, and STL-10~\citep{coates2011analysis}, using multiple model architectures: ResNet-18~\citep{resnet}, PreAct ResNet-18~\citep{he2016identity}, and  WideResNet-28-10~\citep{zagoruyko2016wide}. Results for ResNet-18 are presented in the main paper, while additional evaluations using the other architectures are included in the Appendix (Section~\ref{different_arch}).

CIFAR-10 and CIFAR-100~\citep{krizhevsky2009learning} consist of 60,000 color images (32×32 resolution), split into 50,000 training and 10,000 test samples. CIFAR-10 spans 10 categories, while CIFAR-100 comprises 100 classes. STL-10~\citep{coates2011analysis} includes 5,000 labeled training images and 8,000 test images across 10 classes, originally at 96×96 resolution. For computational efficiency, we resize STL-10 images to 64×64 during training. We trained our approach vs baselines using TRADES as adversarial baseline algorithm. We train our approach with architectures for 150 epochs and  we set learning rate to 0.1, implementing a decay factor of 0.1 for every 5 epochs after 140 epochs. For optimization we used SDG with weight decay factor of 5e-4 and momentunm set to 0.9. At inference, we used adaptive AutoAttack~\citep{lorenz2022is} with pertubation margin $\epsilon = 8/255$ for evaluation of robust performance. Furthermore, we employed a metric inspired from DAFA~\citep{lee2024dafa}, $\varrho$ that is proposed by \citep{li2023wat} to measure the adversarial fairness of our approach vs baselines:

\begin{equation}
\varrho(f, \Delta, \mathcal{A}) = 
\left( \frac{\text{Acc}_{\text{min}}(\mathcal{A}_\Delta(f))}{\text{Acc}_{\text{min}}(\mathcal{A}(f))} 
- 1 \right) - 
\left( \frac{\text{Acc}_{\text{avg}}(\mathcal{A}_\Delta(f))}{\text{Acc}_{\text{avg}}(\mathcal{A}(f))} 
- 1 \right)
\end{equation}

Here, $\text{Acc}_{\text{avg}}(\cdot)$ denotes the average accuracy, while $\text{Acc}_{\text{min}}(\cdot)$ refers to the accuracy of the worst-performing class. The function $\mathcal{A}(f)$ represents a model $f$ trained using a baseline training scheme, whereas $\mathcal{A}_\Delta(f)$ denotes the same model trained using an alternative method incorporating the strategy $\Delta$ to improve fairness. The fairness improvement score $\varrho$ quantifies the relative gain in worst-class performance against the potential loss in overall accuracy. A higher value of $\varrho$ indicates that the method improves fairness (worst-class accuracy) relative to average performance, without significantly sacrificing average performance. To ensure statistical reliability, each method was trained using five random seeds, and results are reported as the average over the final five training epochs. All baselines were evaluated using their official implementations.

For hyperparameter selection, we followed the official settings provided for CIFAR-10 in each method’s original code. For datasets other than CIFAR-10, we experimented with both the default and alternative settings, including those reported in the respective papers and reported the best obtained. Specifically, FRL~\citep{fairness_frl} involves a 40-epoch fine-tuning stage after pretraining; WAT used the 75-90-100 learning rate decay schedule, consistent with TRADES~\citep{defense_trades}; CFA~\citep{wei2023cfa} and BAT adopted a 100-150-200 schedule, following the setup of PGD~\citep{pgd}. All experiments were conducted using a single NVIDIA RTX A6000 GPU with CUDA 12.8 and cuDNN 8.6. We reported results corresponding to the highest worst-class robust accuracy observed across final checkpoints. We set the trade-off parameter $\beta = 6$, which controls the balance between standard accuracy and adversarial robustness, following prior work on TRADES~\citep{defense_trades,lee2024dafa}. The weight adjustment strength $\lambda$, which governs class-wise fairness emphasis, is chosen as $1.0$ for CIFAR-10 and $1.5$ for other datasets, in line with \citep{lee2024dafa}. Training a ResNet-18 model with our proposed approach takes approximately 2.5 hours per random seed on a single GPU, demonstrating its practical efficiency.

\subsection{Additional Experiments} \label{different_arch}

In this section, we provide the additional experiments considering different architectures on different datasets to evaluate the efficacy of our approach vs baselines.

\subsubsection{Experiments on Different Architectures}
We evaluated our approach on PreActResNet-18 and WideResNet-28-10 to showcase the efficacy of our approach on varied architectures. We consider evaluations on CIFAR-10, CIFAR-100 and STL-10 datasets. Table~\ref{tab:evaluations_prn} and Table~\ref{tab:evaluations_wrn} present AutoAttack evaluations of PreActResNet-18 and WideResNet-28-10 models trained with different adversarial fairness approaches on CIFAR-10, CIFAR-100, and STL-10 (Only PreActResNet-18) using a perturbation budget of $\epsilon = 8/255$. We compare TRADES with fairness-aware baselines: DAFA~\citep{lee2024dafa}, FRL~\citep{fairness_frl}, BAT~\citep{fairness_bat}, WAT~\citep{fairness_wat}, CFA~\citep{wei2023cfa}, and FAIR-TAT~\citep{medi2025fair}. We report both average and worst-class accuracies on clean and adversarial samples to assess the fairness-robustness trade-off.

On CIFAR-10 (Table~\ref{tab:cifar10_prn} and~\ref{tab:cifar10_wrn}), most methods improve worst-class clean accuracy over TRADES. BAT achieves the highest average clean accuracy (87.20\%) and worst clean accuracy (74.05\%) with WideResNet-28-10, but suffers from poor adversarial robustness and fairness. FRL also improves fairness on clean samples but at the cost of robust accuracy. TRIX stands out by consistently improving both worst-case clean accuracy and worst-case robust accuracy. On PreActResNet-18, TRIX achieves the highest worst robust accuracy (30.98\%) while maintaining a strong average performance (49.15\%). On WideResNet-28-10, TRIX further improves the robust worst-class accuracy to 34.56\% while keeping the clean performance competitive, showcasing a favorable robustness-fairness trade-off.

Tables~\ref{tab:cifar100_prn} and~\ref{tab:cifar100_wrn} show results on CIFAR-100, a dataset with more classes. Here, TRIX consistently outperforms all baselines in worst-case robust accuracy (2.67\% on PreActResNet-18 and 2.13\% on WideResNet-28-10), while preserving high average robustness. These improvements are notable given the difficulty of achieving fairness in large number of class settings.

On STL-10 (Table~\ref{tab:stl10_prn}), which features higher-resolution images, TRIX again performs best in terms of worst-case robust accuracy (10.08\%) and worst clean accuracy (43.49\%), surpassing both TRADES and other fairness baselines on PreActResNet-18. These results highlight the effectiveness of TRIX in enhancing adversarial fairness across datasets with varying complexity and data resolution, while maintaining overall robustness performance. On STL-10, TRIX further improves the average clean accuracy along with the worst clean accuracy compared to other baselines. In all the evaluations of Tables ~\ref{tab:evaluations_prn} and ~\ref{tab:evaluations_wrn}, the fairness improvement score $\varrho$ is high for our approach in most cases, indicating better balance between the adversarial fairness and robustness.

\begin{table}[t]
\centering
\small
\setlength{\tabcolsep}{5pt}
\caption{AutoAttack evaluation on PreActResNet-18 for different robust fairness approaches with $\epsilon=8/255$. }
\begin{subtable}{\textwidth}
\centering
\caption{CIFAR-10}
\resizebox{\textwidth}{!}{
\begin{tabular}{l|ccc|ccc}
\toprule
\textbf{Method} & \textbf{Avg Clean} & \textbf{Worst Clean} & $\varrho_{\text{clean}}$ & \textbf{Avg Robust} & \textbf{Worst Robust} & $\varrho_{\text{robust}}$ \\
\midrule
TRADES              & 82.35 ± 0.09 & 66.49 ± 1.18 & 0.00 & 49.89 ± 0.16 & 23.42 ± 1.32 & 0.00 \\

TRADES + FRL        &  83.45 ± 0.30 & 69.90 ± 1.17 & 0.04 & 46.23 ± 0.21 & 25.81 ± 0.92 & 0.03\\

TRADES + BAT   &  \textbf{86.81 ± 0.09} & \textbf{73.15 ± 1.32}  & \textbf{0.05} & 48.02 ± 0.19 & 20.30 ± 0.64 & -0.09\\

TRADES + WAT  & 81.20 ± 0.29 & 69.50 ± 1.65 & 0.03  & 46.69 ± 0.35 & 28.01 ± 1.21 & 0.13\\

TRADES + CFA & 79.97 ± 0.41 & 67.01 ± 1.04 & -0.02 & \textbf{50.12 ± 0.20} & 27.09 ± 0.09 & 0.15\\

TRADES + FAIRTAT    & 79.80 ± 0.53 & 68.20 ± 0.24 & -0.01 & 48.71 ± 0.34 & 27.65 ± 0.30 & 0.16\\
TRADES + DAFA       & 82.60 ± 0.14 & 69.20  ± 0.98  &  0.04 & 49.53 ± 0.19 & 30.33 ± 0.90 & 0.29\\
TRADES + TRIX       & 82.63 ± 0.21 & 69.50 ± 1.16 & 0.04 & 49.15 ± 0.15 & \textbf{30.98 ± 1.14} & \textbf{0.31} \\
\bottomrule
\end{tabular}
\label{tab:cifar10_prn}
}
\end{subtable}

\vspace{1em}

\begin{subtable}{\textwidth}
\centering
\caption{CIFAR-100}
\resizebox{\textwidth}{!}{
\begin{tabular}{l|ccc|ccc}
\toprule
\textbf{Method} & \textbf{Avg Clean} & \textbf{Worst Clean} & $\varrho_{\text{clean}}$ & \textbf{Avg Robust} & \textbf{Worst Robust} & $\varrho_{\text{robust}}$ \\
\midrule
TRADES            & 58.33 ± 0.22 & 17.51 ± 1.10 & 0.00 & 25.55 ± 0.18 & 1.37 ± 0.85 & 0.00\\
TRADES + FRL      & 57.91 ± 0.20 & 19.12 ± 1.75 & 0.08 & 25.21 ± 0.29 & 1.65 ± 0.58 & 0.19 \\
TRADES + BAT      & \textbf{62.12 ± 0.21} & \textbf{22.88 ± 1.90} & \textbf{0.24} & 23.32 ± 0.24 & 0.31 ± 0.50 & -0.68 \\
TRADES + WAT      & 53.89 ± 0.35 & 17.40 ± 1.94 & -0.06 & 22.51 ± 0.28 & 1.80 ± 0.14 & 0.19 \\
TRADES + CFA      & 57.71 ± 1.18 & 15.51 ± 1.45 & -0.10 & 23.58 ± 0.26 & 1.59 ± 1.48 & 0.08\\
TRADES + FAIRTAT  & 56.63 ± 1.25 & 16.24 ± 1.15 & -0.04 & 23.83 ± 0.21 & 1.60 ± 1.76 & 0.10 \\
TRADES + DAFA     & 58.11 ± 0.12 & 18.89 ± 1.40 & 0.07 & 25.31 ± 0.19 & 2.21 ± 0.40 & 0.60 \\
TRADES + TRIX     & 59.10 ± 0.18 & 20.15 ± 1.31 & 0.14& \textbf{25.55 ± 0.17} & \textbf{2.67 ± 0.48} & \textbf{0.95}\\
\bottomrule
\end{tabular}
\label{tab:cifar100_prn}
}
\end{subtable}

\vspace{1em}
\begin{subtable}{\textwidth}
\centering
\caption{STL-10}
\resizebox{\textwidth}{!}{
\begin{tabular}{l|ccc|ccc}
\toprule
\textbf{Method} & \textbf{Avg Clean} & \textbf{Worst Clean} & $\varrho_{\text{clean}}$ & \textbf{Avg Robust} & \textbf{Worst Robust} & $\varrho_{\text{robust}}$ \\
\midrule
TRADES              & 61.55 ± 0.65 & 38.45 ± 1.82 & 0.00 & 29.81 ± 0.66 & 6.98 ± 1.14 & 0.00 \\
TRADES + FRL        & 57.63 ± 0.23 & 31.24 ± 1.55 & -0.12 & 28.79 ± 0.41 & 7.26 ± 0.36 & 0.01 \\
TRADES + BAT        & 59.61 ± 1.02 & 36.31 ± 2.45 & -0.02& 24.03 ± 1.07 & 3.37 ± 0.90 & -0.32 \\
TRADES + WAT        & 53.77 ± 1.68 & 31.45 ± 1.36 & -0.05 & 26.53 ± 0.35 & 6.23 ± 0.68 & -0.01 \\
TRADES + CFA        & 60.91 ± 0.61 & 38.70 ± 0.91 & 0.00 & \textbf{31.72 ± 0.22} & 6.96 ± 0.34 & -0.06 \\
TRADES + DAFA       & 60.45 ± 0.59 & 42.13 ± 2.30 & 0.07 & 29.73 ± 0.48 & 9.71 ± 1.21 & 0.39 \\
TRADES + TRIX       & \textbf{61.72 ± 0.66} & \textbf{43.49 ± 2.28} & \textbf{0.13} & 29.75 ± 0.37 & \textbf{10.08 ± 1.33} & \textbf{0.44}\\
\bottomrule
\end{tabular}
\label{tab:stl10_prn}
}
\end{subtable}

\label{tab:evaluations_prn}
\end{table}

\begin{table}[t]
\centering
\small
\setlength{\tabcolsep}{5pt}
\caption{AutoAttack evaluation on WideResNet-28-10 for different robust fairness approaches with $\epsilon=8/255$. }
\begin{subtable}{\textwidth}
\centering
\caption{CIFAR-10}
\resizebox{\textwidth}{!}{
\begin{tabular}{l|ccc|ccc}
\toprule
\textbf{Method} & \textbf{Avg Clean} & \textbf{Worst Clean} & $\varrho_{\text{clean}}$ & \textbf{Avg Robust} & \textbf{Worst Robust} & $\varrho_{\text{robust}}$ \\
\midrule
TRADES              & 82.80 ± 0.10 & 67.10 ± 1.15 & 0.00 & 50.12 ± 0.17 & 24.10 ± 1.30 & 0.00 \\

TRADES + FRL        & 83.60 ± 0.28 & 70.85 ± 1.12 & 0.05& 47.21 ± 0.23 & 27.52 ± 0.90 & 0.08\\

TRADES + BAT        & \textbf{87.20 ± 0.10} & \textbf{74.05 ± 1.25} & \textbf{0.05} & 47.55 ± 0.18 & 20.00 ± 1.33 & -0.12 \\

TRADES + WAT        & 82.30 ± 0.27 & 71.21 ± 1.60 & 0.06 & 48.10 ± 0.33 & 29.81 ± 1.18 & 0.20 \\

TRADES + CFA        & 81.20 ± 0.38 & 66.12 ± 1.02 & 0.00 & 51.20 ± 0.21 & 27.91 ± 0.11 & 0.14\\

TRADES + FAIRTAT    & 81.05 ± 0.50 & 69.11 ± 0.22 & 0.01 & 50.42 ± 0.32 & 29.31 ± 0.28 & 0.21 \\

TRADES + DAFA       & 83.54 ± 0.43 & 70.95 ± 0.87 & 0.05 & 51.86 ± 0.25 & 33.19 ± 1.39 & 0.34 \\

TRADES + TRIX       & 84.98 ± 0.73 & 71.35 ± 0.09 & 0.04 & \textbf{52.13 ± 0.35} & \textbf{34.56 ± 1.32} & \textbf{0.39}\\
\bottomrule
\end{tabular}
\label{tab:cifar10_wrn}
}
\end{subtable}

\vspace{0.5em}

\begin{subtable}{\textwidth}
\centering
\caption{CIFAR-100}
\resizebox{\textwidth}{!}{
\begin{tabular}{l|ccc|ccc}
\toprule
\textbf{Method} & \textbf{Avg Clean} & \textbf{Worst Clean} & $\varrho_{\text{clean}}$ & \textbf{Avg Robust} & \textbf{Worst Robust} & $\varrho_{\text{robust}}$ \\
\midrule
TRADES            & 58.45 ± 0.21 & 17.80 ± 1.10 & 0.00 & 25.76 ± 0.17 & 1.57 ± 0.86 & 0.00\\

TRADES + FRL      & 58.21 ± 0.22 & 19.31 ± 1.87 & 0.08 & 25.29 ± 0.28 & 1.85 ± 0.61 & 0.16 \\

TRADES + BAT      & \textbf{62.40 ± 0.20} & \textbf{22.79 ± 1.93} & \textbf{0.21} & 23.73 ± 0.21 & 0.44 ± 0.59 & -0.64 \\

TRADES + WAT      & 54.02 ± 0.35 & 17.62 ± 2.00 & -0.07 & 22.60 ± 0.25 & 1.83 ± 0.13 & 0.04 \\

TRADES + CFA      & 58.15 ± 1.19 & 15.85 ± 1.42 & -0.10 & 23.79 ± 0.26 & 1.75 ± 1.55 & 0.04 \\

TRADES + DAFA     & 62.30 ± 1.41 & 20.96 ± 1.65 & 0.11 & 27.59 ± 0.54 & 1.89 ± 0.39 & 0.13 \\

TRADES + TRIX     & 63.86 ± 0.23 & 22.38 ± 1.78 & 0.16 & \textbf{27.98 ± 0.45} & \textbf{2.13 ± 0.34} & \textbf{0.27} \\
\bottomrule
\end{tabular}
\label{tab:cifar100_wrn}
}
\end{subtable}

\vspace{1em}

\label{tab:evaluations_wrn}
\end{table}

\subsubsection{Experiments on Tiny-ImageNet} \label{tiny-imagenet}

To further assess the robustness and generalization of our method, we evaluate it on the Tiny-ImageNet dataset~\cite{tinyimagenet}, a 200-class subset of ImageNet with $64 \times 64$ images and 500 training, 50 validation, and 50 test samples per class. Due to fine-grained classes, Tiny-ImageNet serves as a challenging benchmark for evaluating robust performance under limited features across many categories.

Tables ~\ref{tab:tinyimagenet_res}, \ref{tab:tinyimagenet_prn} showcases the evaluations of AutoAttack on Tiny-Imagenet using our approach vs baseline TRADES and DAFA using three random seeds with average over final five checkpoints. We report minimum 20\% classes in case of worst class accuracy as class-wise accuracies are very small. The evaluations prove that our approach achieves better overall robustness and also fairness when provided with dataset like Tiny-Imagenet with limited features per class.

\begin{table}[t]
\centering
\setlength{\tabcolsep}{5pt}
\caption{AutoAttack evaluation on Tiny-ImageNet for different robust fairness approaches with $\epsilon=8/255$. }
\begin{subtable}{\textwidth}
\centering
\caption{ResNet-18}
\resizebox{\textwidth}{!}{
\begin{tabular}{l|ccc|ccc}
\toprule
\textbf{Method} & \textbf{Avg Clean} & \textbf{Worst Clean (Min 20\%)} & $\varrho_{\text{clean}}$ & \textbf{Avg Robust} & \textbf{Worst Robust (Min 20\%)} & $\varrho_{\text{robust}}$ \\
\midrule
TRADES          & 39.87 ± 0.27 & 6.72 ± 1.19 &  0.00 & 11.09 ± 0.25 & 1.19 ± 1.10 & 0.00 \\
TRADES+DAFA     & 39.42 ± 0.31 & 6.56 ± 1.24 &  0.01 & 10.34 ± 0.23 & 2.20 ± 0.35 & 0.78\\
TRADES + TRIX   & \textbf{41.17 ± 0.26} & \textbf{7.34 ± 1.08} &  \textbf{0.06} & \textbf{12.92 ± 0.21} & \textbf{2.96 ± 0.45} & \textbf{1.32} \\
\bottomrule
\end{tabular}
\label{tab:tinyimagenet_res}
}
\end{subtable}
\vspace{0.5em}
\begin{subtable}{\textwidth}
\centering
\caption{PreActResNet-18}
\resizebox{\textwidth}{!}{
\begin{tabular}{l|ccc|ccc}
\toprule
\textbf{Method} & \textbf{Avg Clean} & \textbf{Worst Clean (Min 20\%)} & $\varrho_{\text{clean}}$ & \textbf{Avg Robust} & \textbf{Worst Robust (Min 20\%)} & $\varrho_{\text{robust}}$ \\
\midrule
TRADES           & 39.78 ± 0.47 & 6.06 ± 1.98 & 0.00 & 11.79 ± 0.39 & 1.96 ± 0.52 & 0.00 \\
TRADES + DAFA    & 39.80 ± 0.48 & 6.24 ± 2.05 & 0.03  & 12.44 ± 0.42 & 2.13 ± 0.48 & 0.03 \\
TRADES + TRIX    & \textbf{42.04 ± 0.51} & \textbf{7.42 ± 1.94} & \textbf{0.12} & \textbf{12.74 ± 0.35} & \textbf{2.32 ± 0.44} & \textbf{0.10} \\
\bottomrule
\end{tabular}
\label{tab:tinyimagenet_prn}
}
\end{subtable}
\end{table}

\subsection{Ablation Studies} \label{ablation}

\subsubsection{Ablations with Different Data Augmentations}

Table~\ref{tab:augmentation_ablations} evaluates the impact of different data augmentation strategies like AutoAugment~\cite{cubuk2019autoaugment}, Cutout~\cite{devries2017cutout}, and Mixup~\cite{zhang2018mixup} on the TRIX and recent baseline DAFA frameworks under AutoAttack on CIFAR-10 using ResNet-18. All augmentations yield notable gains over the TRADES baseline, particularly in worst-case robust accuracy.

TRIX benefits most from augmentations, with \textbf{TRIX + Mixup} achieving the best Worst Robust accuracy (32.6\%) and highest Avg Robust accuracy (51.0\%). DAFA also shows consistent improvements across augmentations. These results suggest that augmentations increase the model’s effective representation capacity, leading to better generalization and robustness under adversarial settings. This aligns with the hypothesis that richer augmentations encourage more diverse and discriminative features, which help models perform better in both clean and adversarial regimes.

\begin{table}[ht]
\centering
\caption{AutoAttack results on CIFAR-10 with ResNet-18 ($\epsilon=8/255$), comparing TRIX and DAFA under different data augmentations.}
\scriptsize
\resizebox{\textwidth}{!}{
\begin{tabular}{l|c|c|c|c}
\toprule
\textbf{Method} & \textbf{Avg Clean (\%)} & \textbf{Worst Clean (\%)} & \textbf{Avg Robust (\%)} & \textbf{Worst Robust (\%)} \\
\midrule
TRADES & 82.0 & 65.7 & 49.8 & 23.1 \\
TRIX + AutoAug & \textbf{83.6} & \textbf{71.9} & \textbf{50.3} & \textbf{32.1} \\
TRIX + Cutout  & \textbf{83.1} & \textbf{71.2} & \textbf{50.6} & \textbf{31.9} \\
TRIX + Mixup   & \textbf{82.6} & \textbf{70.1} & \textbf{51.0} & \textbf{32.6} \\
DAFA + AutoAug & 82.1 & 69.8 & 50.1 & 31.4 \\
DAFA + Cutout  & 81.7 & 70.1 & 50.5 & 31.5 \\
DAFA + Mixup   & 81.9 & 70.3 & 50.6 & 32.1 \\
\bottomrule
\end{tabular}
}
\label{tab:augmentation_ablations}
\end{table}

\vspace{0.5em}
\subsection{Theoretical Analysis} \label{Theoretical}

In this section, we provide a theoretical justification for the design of \textbf{TRIX}, a mixed adversarial training framework. TRIX applies \emph{untargeted adversaries} to \emph{vulnerable (weak)} classes and \emph{targeted adversaries} to \emph{strong (well-separated)} classes. This class-aware strategy is rooted in classical multi-class decomposition schemes as combination of One-vs-All and One-vs-One classifier scheme \citep{PAWARA2020107528} as follows:

\begin{itemize}
    \item The \textbf{One-vs-All (OvA)} classifier scheme resembles defense against \emph{untargeted adversaries}, which push a class away from all others simultaneously. This is well suited for weak classes that are often confused with multiple others.
    
    \item The \textbf{One-vs-One (OvO)} classifier scheme mimics the defense against \emph{targeted adversaries}, which separate one class from another specific class. This is more efficient for strong classes that already have clear separation.
\end{itemize}

TRIX leverages this structure: untargeted (OvA-like) objectives emphasize confused regions of the feature space, while targeted (OvO-like) objectives refine already-separated regions. Crucially, TRIX also \textbf{assigns higher loss weights to weak classes}, ensuring they receive more attention during training to preserve fairness and robustness in fragile parts of the space.

\begin{theorem}[OvO vs. OvA Weight Complexity and TRIX Motivation] \label{thm:ovo_ova}
Let $\{h_i\}_{i=1}^K$ be $K$ class-representative adversarial examples in $\mathbb{R}$ such that all class pairs $(i,j)$ are linearly separable. Then:

\begin{enumerate}
    \item Any One-vs-One (OvO) classifier trained on class pair $(i,j)$ with margin $R > 0$ requires weight magnitude of order $\mathcal{O}(1/\delta_{ij})$, where $\delta_{ij} = |h_i - h_j|$.
    
    \item Any One-vs-All (OvA) classifier for class $i$, separating it from all other classes with margin $R$, requires weight magnitude of order $\mathcal{O}(KR)$ under uniform spacing $h_i = i - 1$.
    
    \item Consequently, TRIX's use of OvA-style untargeted losses for weak classes, paired with higher weighting, ensures broad separation and fairness. Meanwhile, OvO-style targeted losses for strong classes reduce margin errors without excessive capacity cost.
\end{enumerate}
\end{theorem}

We analyze both schemes assuming each linear classifier is of the form $f(h) = wh + b$, and margin $R > 0$ must be maintained between class boundaries.

\paragraph{1. OvO Scheme.}  
For a pair $(i,j)$, a classifier $f_{ij}(h) = w_{ij} h + b_{ij}$ must satisfy:
\[
f_{ij}(h_i) \ge R, \quad f_{ij}(h_j) \le -R
\Rightarrow w_{ij}(h_i - h_j) \ge 2R \Rightarrow |w_{ij}| \ge \frac{2R}{\delta_{ij}}
\]
This weight depends only on pairwise distance and remains bounded:
\[
|w_{ij}| = \mathcal{O}\left(\frac{1}{\delta_{ij}}\right)
\]

\paragraph{2. OvA Scheme.}  
A classifier $f_i(h) = w_i h + b_i$ must separate class $i$ from all others:
\[
f_i(h_i) \ge f_i(h_j) + R \quad \forall j \ne i
\Rightarrow w_i(h_i - h_j) \ge R
\]
Assuming equidistant spacing $h_i = i - 1$, the tightest constraint arises from the closest class:
\[
|h_i - h_j| = \mathcal{O}(1) \Rightarrow |w_i| \ge \mathcal{O}(R)
\]
But to satisfy all $K - 1$ inequalities simultaneously, $w_i$ must satisfy the worst-case gap:
\[
|w_i| = \mathcal{O}(KR)
\]

\paragraph{3. Implication for TRIX.}  
This reveals that OvA-style losses, while effective at global separation, are more costly in terms of required weight norms. TRIX therefore uses:

\begin{itemize}
    \item \textbf{OvA-style (untargeted) adversarial training with higher weights} for \textit{weak or confused classes}, ensuring they are pushed away from all others and receive more training attention to correct systemic bias or collapse.
    
    \item \textbf{OvO-style (targeted) adversarial training with lower weights} for \textit{strong classes}, since they only require refinement against a few confusing neighbors.
\end{itemize}

This class-aware weighting is essential to preserving \emph{fairness}, as it prevents underperforming classes from being underrepresented in gradient updates.

\paragraph{Conclusion.}  
Theorem~\ref{thm:ovo_ova} formalizes the key motivation behind TRIX: untargeted attacks (OvA) are expansive and weight expensive; targeted attacks (OvO) are efficient but narrow. By combining both, TRIX achieves balanced robustness and fairness across all classes, directing more attention to fragile regions. 
\\
Code of TRIX can be found at \url{https://github.com/Fairness-a11y/TRIX/tree/main}.

\subsection{Feature Space Coverage}
\newlength{\imgwidth}
\setlength{\imgwidth}{0.35\textwidth}
\begin{figure}[H]
    \centering
    \begin{tabular}{c|c}
        \textbf{Strong Classes} & \textbf{Weak Classes} \\
        \midrule
        \includegraphics[width=\imgwidth]{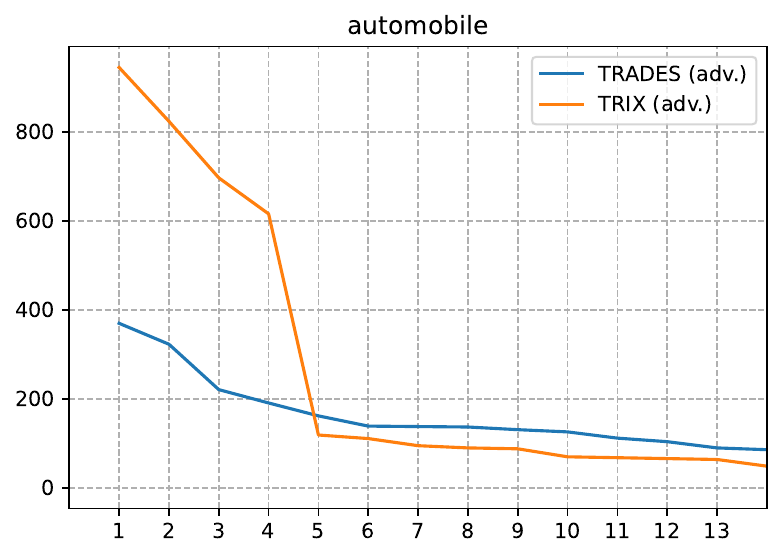} &
        \includegraphics[width=\imgwidth]{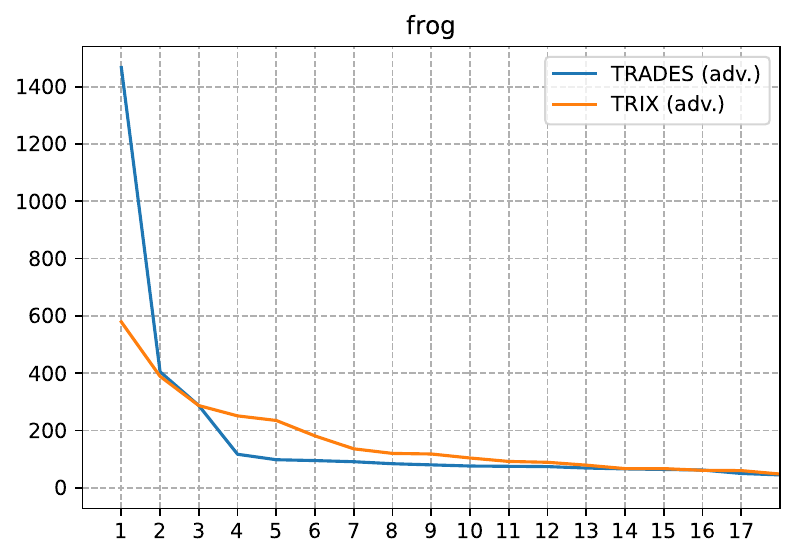} \\
        \includegraphics[width=\imgwidth]{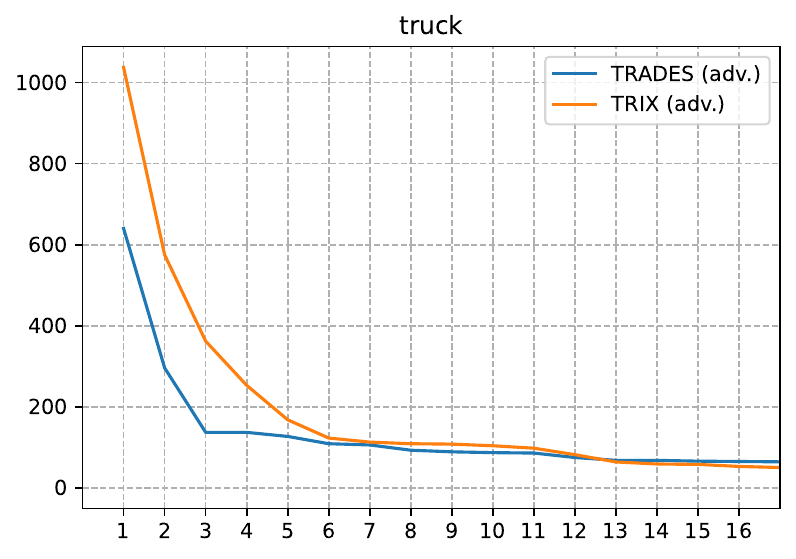} &
        \includegraphics[width=\imgwidth]{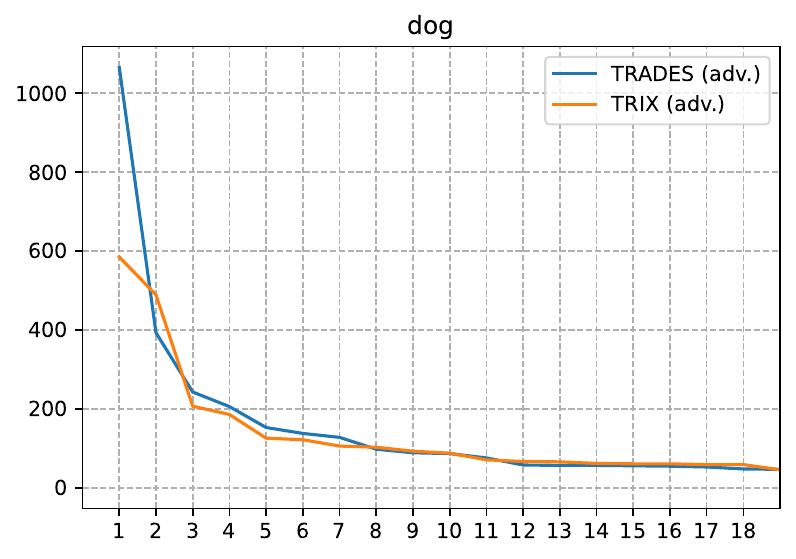} \\
        \includegraphics[width=\imgwidth]{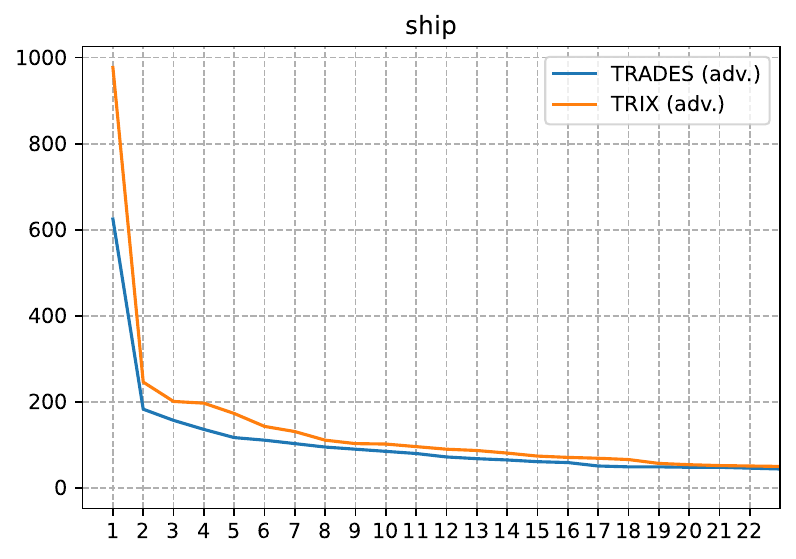} &
        \includegraphics[width=\imgwidth]{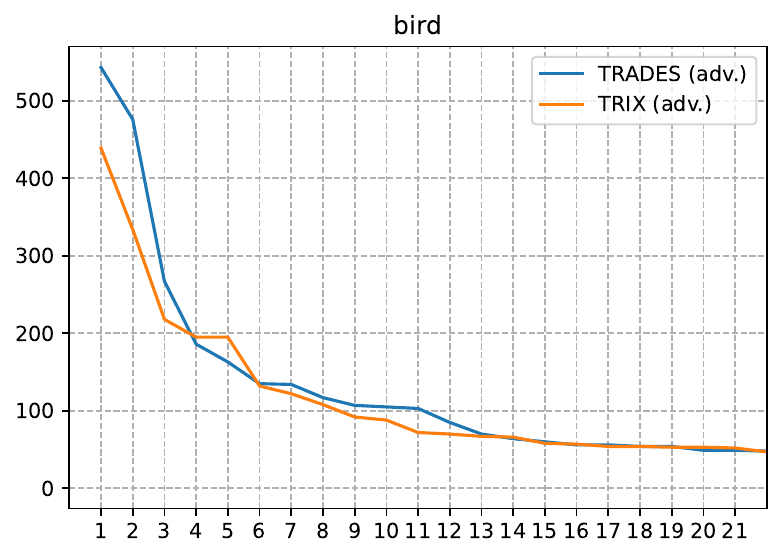} \\
        \includegraphics[width=\imgwidth]{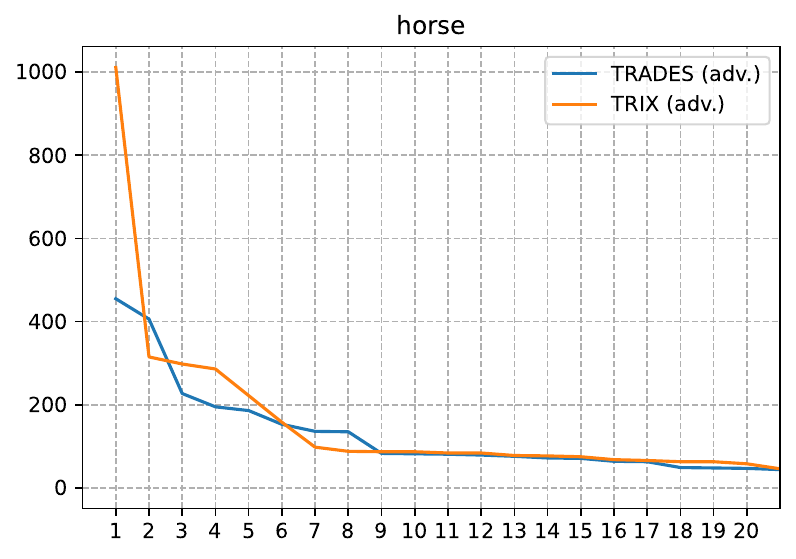} &
        \includegraphics[width=\imgwidth]{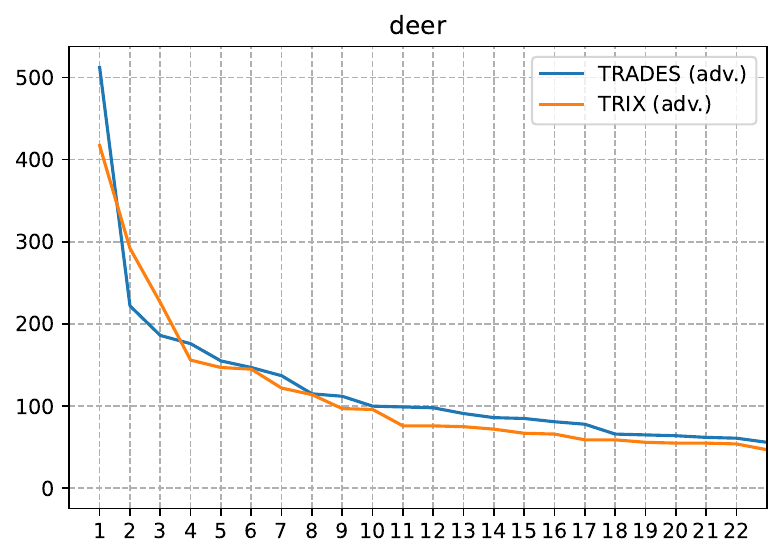} \\
        \includegraphics[width=\imgwidth]{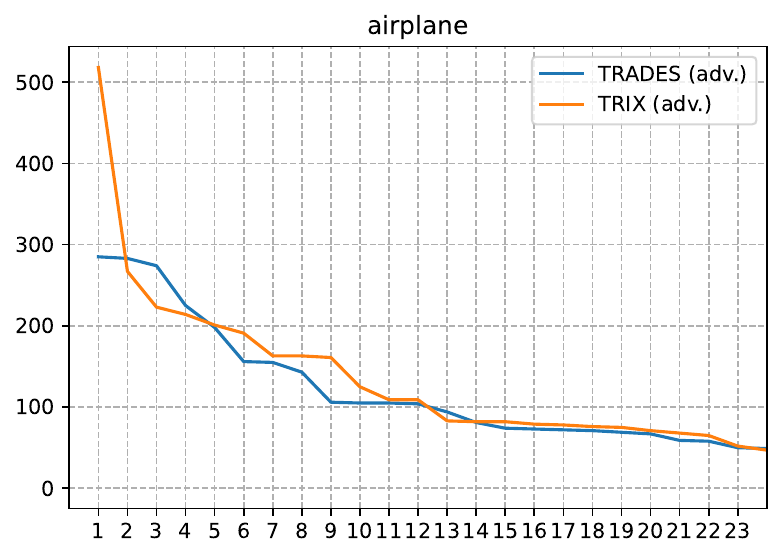} &
        \includegraphics[width=\imgwidth]{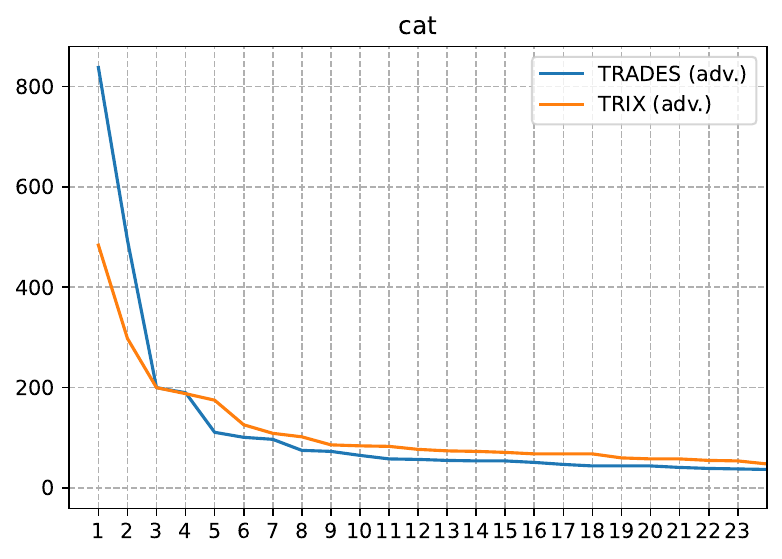} \\
    \end{tabular}
    \caption{%
    We measure the coverage of the feature space of each class by projecting class-wise image features onto the unit sphere and assigning them to $1M$ randomly sampled bins.}
    \label{fig:supp:feature-coverage}
\end{figure}

{\small
\bibliographystyle{unsrt}
\bibliography{egbib}

\begin{thebibliography}{10}

\bibitem{he2016deep}
Kaiming He, Xiangyu Zhang, Shaoqing Ren, and Jian Sun.
\newblock Deep residual learning for image recognition.
\newblock In {\em IEEE Conference on Computer Vision and Pattern Recognition (CVPR)}, pages 770--778, 2016.

\bibitem{dosovitskiy2021image}
Alexey Dosovitskiy, Lucas Beyer, Alexander Kolesnikov, Dirk Weissenborn, Xiaohua Zhai, Thomas Unterthiner, et~al.
\newblock An image is worth 16x16 words: Transformers for image recognition at scale.
\newblock In {\em International Conference on Learning Representations (ICLR)}, 2021.

\bibitem{devlin2019bert}
Jacob Devlin, Ming-Wei Chang, Kenton Lee, and Kristina Toutanova.
\newblock Bert: Pre-training of deep bidirectional transformers for language understanding.
\newblock In {\em North American Chapter of the Association for Computational Linguistics (NAACL)}, pages 4171--4186, 2019.

\bibitem{brown2020language}
Tom~B. Brown, Benjamin Mann, Nick Ryder, Melanie Subbiah, Jared Kaplan, Prafulla Dhariwal, et~al.
\newblock Language models are few-shot learners.
\newblock In {\em Advances in Neural Information Processing Systems (NeurIPS)}, pages 1877--1901, 2020.

\bibitem{silver2017mastering}
David Silver, Julian Schrittwieser, Karen Simonyan, et~al.
\newblock Mastering the game of go without human knowledge.
\newblock {\em Nature}, 550(7676):354--359, 2017.

\bibitem{szegedy2014intriguing}
Christian Szegedy, Wojciech Zaremba, Ilya Sutskever, Joan Bruna, Dumitru Erhan, Ian Goodfellow, and Rob Fergus.
\newblock Intriguing properties of neural networks.
\newblock In {\em International Conference on Learning Representations (ICLR)}, 2014.

\bibitem{goodfellow2014explaining}
Ian~J Goodfellow, Jonathon Shlens, and Christian Szegedy.
\newblock Explaining and harnessing adversarial examples.
\newblock In {\em International Conference on Learning Representations (ICLR)}, 2015.

\bibitem{madry2018towards}
Aleksander Madry, Aleksandar Makelov, Ludwig Schmidt, Dimitris Tsipras, and Adrian Vladu.
\newblock Towards deep learning models resistant to adversarial attacks.
\newblock In {\em International Conference on Learning Representations (ICLR)}, 2018.

\bibitem{kurakin2017adversarial}
Alexey Kurakin, Ian Goodfellow, and Samy Bengio.
\newblock Adversarial examples in the physical world.
\newblock In {\em International Conference on Learning Representations (ICLR) Workshop}, 2017.

\bibitem{carlini2017towards}
Nicholas Carlini and David Wagner.
\newblock Towards evaluating the robustness of neural networks.
\newblock In {\em IEEE Symposium on Security and Privacy}, pages 39--57, 2017.

\bibitem{agnihotri2025sembench}
Shashank Agnihotri, David Schader, Jonas Jakubassa, Nico Sharei, Simon Kral, Mehmet~Ege Kaçar, Ruben Weber, and Margret Keuper.
\newblock Semsegbench \& detecbench: Benchmarking reliability and generalization beyond classification, 2025.

\bibitem{athalye2018obfuscated}
Anish Athalye, Nicholas Carlini, and David Wagner.
\newblock Obfuscated gradients give a false sense of security: Circumventing defenses to adversarial examples.
\newblock In {\em International Conference on Machine Learning (ICML)}, pages 274--283, 2018.

\bibitem{Wang2020ImprovingAR}
Yisen Wang, Difan Zou, Jinfeng Yi, James Bailey, Xingjun Ma, and Quanquan Gu.
\newblock Improving adversarial robustness requires revisiting misclassified examples.
\newblock In {\em International Conference on Learning Representations}, 2020.

\bibitem{jia2022prior}
Xiaojun Jia, Yong Zhang, Xingxing Wei, Baoyuan Wu, Ke~Ma, Jue Wang, and Xiaochun Cao.
\newblock Prior-guided adversarial initialization for fast adversarial training.
\newblock In {\em European Conference on Computer Vision}, pages 567--584. Springer, 2022.

\bibitem{10.1007/978-3-031-73636-0_21}
Shashank Agnihotri, Julia Grabinski, and Margret Keuper.
\newblock Improving feature stability during upsampling -- spectral artifacts and the importance of spatial context.
\newblock In Ale{\v{s}} Leonardis, Elisa Ricci, Stefan Roth, Olga Russakovsky, Torsten Sattler, and G{\"u}l Varol, editors, {\em Computer Vision -- ECCV 2024}, pages 357--376, Cham, 2025. Springer Nature Switzerland.

\bibitem{agnihotri2023cospgd}
Shashank Agnihotri, Steffen Jung, and Margret Keuper.
\newblock Cospgd: a unified white-box adversarial attack for pixel-wise prediction tasks.
\newblock {\em arXiv preprint arXiv:2302.02213}, 2023.

\bibitem{agnihotri2023unreasonable}
Shashank Agnihotri, Kanchana~Vaishnavi Gandikota, Julia Grabinski, Paramanand Chandramouli, and Margret Keuper.
\newblock On the unreasonable vulnerability of transformers for image restoration-and an easy fix.
\newblock In {\em Proceedings of the IEEE/CVF International Conference on Computer Vision}, pages 3707--3717, 2023.

\bibitem{grabinski2024large}
Julia Grabinski, Janis Keuper, and Margret Keuper.
\newblock As large as it gets-studying infinitely large convolutions via neural implicit frequency filters.
\newblock {\em Transactions on Machine Learning Research}, 2024:1--42, 2024.

\bibitem{Grabinskilowcut22}
Julia Grabinski, Steffen Jung, Janis Keuper, and Margret Keuper.
\newblock Frequencylowcut pooling - plug and play against catastrophic overfitting.
\newblock In {\em Computer Vision - {ECCV} 2022 - 17th European Conference, Tel Aviv, Israel, October 23-27, 2022, Proceedings, Part {XIV}}, pages 36--57. Springer, 2022.

\bibitem{grabinski2022aliasing}
Julia Grabinski, Janis Keuper, and Margret Keuper.
\newblock Aliasing and adversarial robust generalization of cnns.
\newblock {\em Machine Learning}, 111(11):3925--3951, 2022.

\bibitem{grabinski2022robust}
Julia Grabinski, Paul Gavrikov, Janis Keuper, and Margret Keuper.
\newblock Robust models are less over-confident.
\newblock {\em Advances in Neural Information Processing Systems}, 35:39059--39075, 2022.

\bibitem{Jung2023}
Steffen Jung, Jovita Lukasik, and Margret Keuper.
\newblock Neural architecture design and robustness: A dataset.
\newblock In {\em ICLR}, 2023.

\bibitem{lukasik2023improving}
Jovita Lukasik, Paul Gavrikov, Janis Keuper, and Margret Keuper.
\newblock Improving native cnn robustness with filter frequency regularization.
\newblock {\em Transactions on Machine Learning Research}, 2023:1--36, 2023.

\bibitem{zhang2019theoretically}
Hongyang Zhang, Yaodong Yu, Jiantao Jiao, Eric~P Xing, Laurent El~Ghaoui, and Michael~I Jordan.
\newblock Theoretically principled trade-off between robustness and accuracy.
\newblock In {\em International Conference on Machine Learning (ICML)}, pages 7472--7482, 2019.

\bibitem{tian2021analysis}
Qi~Tian, Kun Kuang, Kelu Jiang, Fei Wu, and Yisen Wang.
\newblock Analysis and applications of class-wise robustness in adversarial training.
\newblock In {\em Proceedings of the 27th ACM SIGKDD Conference on Knowledge Discovery \& Data Mining}, pages 1561--1570, 2021.

\bibitem{xu2021robust}
Depeng Xu, Sen Yuan, Hongying Zhang, and Xintao Wu.
\newblock Robust fairness: A robust optimization framework for fair classification.
\newblock In {\em IEEE International Conference on Data Mining (ICDM)}, pages 721--730, 2021.

\bibitem{zhang2021dafa}
Zhilu Zhang, Tong Xu, Hanghang Zhang, Jundong Wang, and Xia~Hu Huang.
\newblock Dafa: Differentiated adversarial training for fairness and accuracy.
\newblock In {\em International Conference on Learning Representations (ICLR)}, 2021.

\bibitem{medi2024classwiserobustnessanalysis}
Tejaswini Medi, Julia Grabinski, and Margret Keuper.
\newblock Towards class-wise robustness analysis, 2024.

\bibitem{medi2025fair}
Tejaswini Medi, Steffen Jung, and Margret Keuper.
\newblock Fair-tat: Improving model fairness using targeted adversarial training.
\newblock In {\em 2025 IEEE/CVF Winter Conference on Applications of Computer Vision (WACV)}, pages 7827--7836. IEEE, 2025.

\bibitem{wei2023cfa}
Zeming Wei, Yifei Wang, Yiwen Guo, and Yisen Wang.
\newblock Cfa: Class-wise calibrated fair adversarial training.
\newblock In {\em Proceedings of the IEEE/CVF Conference on Computer Vision and Pattern Recognition}, pages 8193--8201, 2023.

\bibitem{fairness_bat}
A.~Gupta and L.~Tan.
\newblock Decomposing fairness in adversarial settings: Source vs. target bias.
\newblock In {\em International Conference on Machine Learning (ICML)}, 2023.

\bibitem{li2023wat}
Boqi Li and Weiwei Liu.
\newblock Wat: improve the worst-class robustness in adversarial training.
\newblock In {\em Proceedings of the AAAI conference on artificial intelligence}, volume~37, pages 14982--14990, 2023.

\bibitem{fairness_frl}
T.~Liu and H.~Zhao.
\newblock Theoretical analysis of class-wise risk in adversarial training.
\newblock In {\em Advances in Neural Information Processing Systems (NeurIPS)}, 2022.

\bibitem{ma2021understanding}
Xingjun Ma, Yuhao Niu, Lin Gu, Yisen Wang, Yitian Zhao, James Bailey, and Feng Lu.
\newblock Understanding adversarial attacks on deep learning based medical image analysis systems.
\newblock {\em Pattern Recognition}, 110:107332, 2021.

\bibitem{lee2024dafa}
Hyungyu Lee, Saehyung Lee, Hyemi Jang, Junsung Park, Ho~Bae, and Sungroh Yoon.
\newblock Dafa: Distance-aware fair adversarial training.
\newblock In {\em 12th International Conference on Learning Representations, ICLR 2024}, 2024.

\bibitem{ma2022tradeoff}
Xinsong Ma, Zekai Wang, and Weiwei Liu.
\newblock On the tradeoff between robustness and fairness.
\newblock {\em Advances in Neural Information Processing Systems}, 35:26230--26241, 2022.

\bibitem{zhao2023improving}
Shiji Zhao, Xizhe Wang, and Xingxing Wei.
\newblock Improving adversarial robust fairness via anti-bias soft label distillation.
\newblock {\em arXiv preprint arXiv:2312.05508}, 2023.

\bibitem{li2024adversarial}
Binghui Li and Yuanzhi Li.
\newblock Adversarial training can provably improve robustness: Theoretical analysis of feature learning process under structured data.
\newblock {\em arXiv preprint arXiv:2410.08503}, 2024.

\bibitem{targeted_adv_training}
Gauri Ding, Yuzhe Liu, Michael~I. Jordan, and Jacob Steinhardt.
\newblock Mma training: Direct input space margin maximization through adversarial training.
\newblock In {\em International Conference on Learning Representations (ICLR)}, 2020.

\bibitem{fairness_analysis}
S.~Kumar and R.~Singh.
\newblock On the trade-off between robustness and fairness in adversarial settings.
\newblock In {\em International Conference on Machine Learning (ICML)}, 2022.

\bibitem{fairness_understanding}
Yuzheng Hu, Fan Wu, Hongyang Zhang, and Han Zhao.
\newblock Understanding the impact of adversarial robustness on accuracy disparity.
\newblock In {\em International Conference on Machine Learning}, pages 13679--13709. PMLR, 2023.

\bibitem{fairness_kdd}
Qi~Tian, Kun Kuang, Kelu Jiang, Fei Wu, and Yisen Wang.
\newblock Analysis and applications of class-wise robustness in adversarial training.
\newblock In {\em Proceedings of the 27th ACM SIGKDD Conference on Knowledge Discovery \& Data Mining}, pages 1561--1570, 2021.

\bibitem{fairness_weighting}
Philipp Benz, Chaoning Zhang, Adil Karjauv, and In~So Kweon.
\newblock Robustness may be at odds with fairness: An empirical study on class-wise accuracy.
\newblock In {\em NeurIPS 2020 Workshop on Pre-registration in Machine Learning}, pages 325--342. PMLR, 2021.

\bibitem{xu2021frl}
A.~Xu, B.~Li, and Y.~Wang.
\newblock Towards fair robust learning: Adjusting adversarial margins and weights.
\newblock In {\em Advances in Neural Information Processing Systems (NeurIPS)}, 2021.

\bibitem{ma2021fat}
C.~Ma, Y.~Zhang, and L.~Lyu.
\newblock Fair adversarial training via risk variance regularization.
\newblock In {\em Advances in Neural Information Processing Systems (NeurIPS)}, 2021.

\bibitem{sun2021bat}
Y.~Sun, M.~Long, and J.~Wang.
\newblock Balanced adversarial training for source and target class fairness.
\newblock In {\em International Conference on Learning Representations (ICLR)}, 2021.

\bibitem{wu2021entropy}
X.~Wu, Y.~Zhang, and J.~Zou.
\newblock Entropy regularization improves fairness in adversarial learning.
\newblock In {\em International Conference on Learning Representations (ICLR)}, 2021.

\bibitem{yue2023fairard}
Z.~Yue, Y.~Zhang, and B.~Han.
\newblock Fair-ard: Reweighting based on class vulnerability for robust fairness.
\newblock In {\em International Conference on Machine Learning (ICML)}, 2023.

\bibitem{fairness_wat}
J.~Zhao and Q.~Huang.
\newblock Validation-guided fairness-aware adversarial training.
\newblock In {\em Advances in Neural Information Processing Systems (NeurIPS)}, 2023.

\bibitem{cifar10}
Alex Krizhevsky, Vinod Nair, and Geoffrey Hinton.
\newblock Cifar-10 (canadian institute for advanced research).
\newblock {\em Dataset}, 2009.

\bibitem{coates2011analysis}
Adam Coates, Andrew~Y Ng, and Honglak Lee.
\newblock An analysis of single-layer networks in unsupervised feature learning.
\newblock In {\em Proceedings of the fourteenth international conference on artificial intelligence and statistics}, pages 215--223. JMLR Workshop and Conference Proceedings, 2011.

\bibitem{he_deep_2016}
Kaiming He, Xiangyu Zhang, Shaoqing Ren, and Jian Sun.
\newblock Deep residual learning for image recognition.
\newblock In {\em 2016 {IEEE} Conference on Computer Vision and Pattern Recognition ({CVPR})}, pages 770--778, 2016.
\newblock {ISSN}: 1063-6919.

\bibitem{attack_aaa}
Minhao Huang, Yinpeng Dong, Hang Su, and Jun Zhu.
\newblock Reliable evaluation of adversarial robustness with an ensemble of diverse parameter-free attacks.
\newblock In {\em Advances in Neural Information Processing Systems (NeurIPS)}, 2020.

\bibitem{krizhevsky2009learning}
Alex Krizhevsky, Geoffrey Hinton, et~al.
\newblock Learning multiple layers of features from tiny images.
\newblock {\em https://www.cs.toronto.edu/~kriz/learning-features-2009-TR.pdf}, 2009.

\bibitem{resnet}
Kaiming He, Xiangyu Zhang, Shaoqing Ren, and Jian Sun.
\newblock Deep residual learning for image recognition, 2015.

\bibitem{he2016identity}
Kaiming He, Xiangyu Zhang, Shaoqing Ren, and Jian Sun.
\newblock Identity mappings in deep residual networks.
\newblock In {\em Computer Vision--ECCV 2016: 14th European Conference, Amsterdam, The Netherlands, October 11--14, 2016, Proceedings, Part IV 14}, pages 630--645. Springer, 2016.

\bibitem{zagoruyko2016wide}
Sergey Zagoruyko and Nikos Komodakis.
\newblock Wide residual networks.
\newblock {\em arXiv preprint arXiv:1605.07146}, 2016.

\bibitem{lorenz2022is}
Peter Lorenz, Dominik Strassel, Margret Keuper, and Janis Keuper.
\newblock Is robustbench/autoattack a suitable benchmark for adversarial robustness?
\newblock In {\em The AAAI-22 Workshop on Adversarial Machine Learning and Beyond}, 2022.

\bibitem{defense_trades}
Hongyang Zhang, Yaodong Yu, Jiantao Jiao, Eric Xing, Laurent~El Ghaoui, and Michael~I. Jordan.
\newblock Theoretically principled trade-off between robustness and accuracy.
\newblock In {\em International Conference on Machine Learning (ICML)}, 2019.

\bibitem{pgd}
Alexey Kurakin, Ian Goodfellow, and Samy Bengio.
\newblock Adversarial machine learning at scale, 2017.

\bibitem{tinyimagenet}
Y.~Le and X.~Yang.
\newblock Tiny imagenet visual recognition challenge.
\newblock \url{https://tiny-imagenet.herokuapp.com/}, 2015.

\bibitem{cubuk2019autoaugment}
Ekin~D Cubuk, Barret Zoph, Dandelion Mane, Vijay Vasudevan, and Quoc~V Le.
\newblock Autoaugment: Learning augmentation strategies from data.
\newblock In {\em CVPR}, 2019.

\bibitem{devries2017cutout}
Terrance DeVries and Graham~W Taylor.
\newblock Improved regularization of convolutional neural networks with cutout.
\newblock In {\em arXiv preprint arXiv:1708.04552}, 2017.

\bibitem{zhang2018mixup}
Hongyi Zhang, Moustapha Cisse, Yann~N Dauphin, and David Lopez-Paz.
\newblock mixup: Beyond empirical risk minimization.
\newblock In {\em International Conference on Learning Representations (ICLR)}, 2018.

\bibitem{PAWARA2020107528}
Pornntiwa Pawara, Emmanuel Okafor, Marc Groefsema, Sheng He, Lambert~R.B. Schomaker, and Marco~A. Wiering.
\newblock One-vs-one classification for deep neural networks.
\newblock {\em Pattern Recognition}, 108:107528, 2020.

\end{thebibliography}
}


\newpage
\section*{NeurIPS Paper Checklist}

\begin{enumerate}

\item {\bf Claims}
    \item[] Question: Do the main claims made in the abstract and introduction accurately reflect the paper's contributions and scope?
    \item[] Answer: \answerYes{} 
    \item[] Justification: See Sec.\ref{Intro}.

\item {\bf Limitations}
    \item[] Question: Does the paper discuss the limitations of the work performed by the authors?
    \item[] Answer: \answerYes{} 
    \item[] Justification: See below Sec.\ref{conclusion}.

\item {\bf Theory assumptions and proofs}
    \item[] Question: For each theoretical result, does the paper provide the full set of assumptions and a complete (and correct) proof?
    \item[] Answer: \answerNA{} 

    \item {\bf Experimental result reproducibility}
    \item[] Question: Does the paper fully disclose all the information needed to reproduce the main experimental results of the paper to the extent that it affects the main claims and/or conclusions of the paper (regardless of whether the code and data are provided or not)?
    \item[] Answer: \answerYes{} 
    \item[] Justification: See Sec. \ref{exps} and Appendix \ref{additional_details}.
    
   
\item {\bf Open access to data and code}
    \item[] Question: Does the paper provide open access to the data and code, with sufficient instructions to faithfully reproduce the main experimental results, as described in supplemental material?
    \item[] Answer: \answerYes{} 
    \item[] Justification: The code is available at supplementary material.

\item {\bf Experimental setting/details}
    \item[] Question: Does the paper specify all the training and test details (e.g., data splits, hyperparameters, how they were chosen, type of optimizer, etc.) necessary to understand the results?
    \item[] Answer: \answerYes{} 
     \item[] Justification: See Sec. \ref{exps} and Appendix \ref{additional_details}.

\item {\bf Experiment statistical significance}
    \item[] Question: Does the paper report error bars suitably and correctly defined or other appropriate information about the statistical significance of the experiments?
    \item[] Answer: \answerYes{} 
    \item[] Justification: We include mean and standard deviation.

\item {\bf Experiments compute resources}
    \item[] Question: For each experiment, does the paper provide sufficient information on the computer resources (type of compute workers, memory, time of execution) needed to reproduce the experiments?
    \item[] Answer: \answerYes{} 
    \item[] Justification: See Appendix \ref{additional_details}.
    
\item {\bf Code of ethics}
    \item[] Question: Does the research conducted in the paper conform, in every respect, with the NeurIPS Code of Ethics \url{https://neurips.cc/public/EthicsGuidelines}?
    \item[] Answer: \answerYes{} 

\item {\bf Broader impacts}
    \item[] Question: Does the paper discuss both potential positive societal impacts and negative societal impacts of the work performed?
    \item[] Answer: \answerYes{} 
    \item[] Justification: See Sec.\ref{conclusion}.
    
\item {\bf Safeguards}
    \item[] Question: Does the paper describe safeguards that have been put in place for responsible release of data or models that have a high risk for misuse (e.g., pretrained language models, image generators, or scraped datasets)?
    \item[] Answer: \answerNA{} 

\item {\bf Licenses for existing assets}
    \item[] Question: Are the creators or original owners of assets (e.g., code, data, models), used in the paper, properly credited and are the license and terms of use explicitly mentioned and properly respected?
    \item[] Answer: \answerYes{} 

\item {\bf New assets}
    \item[] Question: Are new assets introduced in the paper well documented and is the documentation provided alongside the assets?
    \item[] Answer: \answerNA{} 

\item {\bf Crowdsourcing and research with human subjects}
    \item[] Question: For crowdsourcing experiments and research with human subjects, does the paper include the full text of instructions given to participants and screenshots, if applicable, as well as details about compensation (if any)? 
    \item[] Answer: \answerNA{} 

\item {\bf Institutional review board (IRB) approvals or equivalent for research with human subjects}
    \item[] Question: Does the paper describe potential risks incurred by study participants, whether such risks were disclosed to the subjects, and whether Institutional Review Board (IRB) approvals (or an equivalent approval/review based on the requirements of your country or institution) were obtained?
    \item[] Answer: \answerNA{} 

\item {\bf Declaration of LLM usage}
    \item[] Question: Does the paper describe the usage of LLMs if it is an important, original, or non-standard component of the core methods in this research? Note that if the LLM is used only for writing, editing, or formatting purposes and does not impact the core methodology, scientific rigorousness, or originality of the research, declaration is not required.
    \item[] Answer: \answerNo{} 
\end{enumerate}

\newpage

\end{document}